\documentclass[11pt]{article}

\usepackage[final]{acl}

\usepackage{times}
\usepackage{latexsym}

\usepackage[T1]{fontenc}

\usepackage[utf8]{inputenc}

\usepackage{microtype}

\usepackage{inconsolata}

\usepackage{graphicx}
\usepackage{booktabs}
\usepackage{amsmath}
\usepackage{multirow}
\usepackage{colortbl}
\usepackage[most]{tcolorbox}
\usepackage[dvipsnames]{xcolor}
\usepackage{xcolor}
\usepackage{algorithm}
\usepackage{algpseudocode}
\usepackage{amsmath}
\usepackage{amssymb}

\definecolor{lightgreen}{rgb}{0.9,1.0,0.9} 
\definecolor{lightpink}{rgb}{1.0,0.92,0.95} 
\newcommand{\revise}[1]{{\color{black} #1}}
\newcommand{\highlight}[1]{{\color{black} #1}}
\newcommand{\camera}[1]{{\color{black} #1}}

\title{Interactive Learning for LLM Reasoning}

\author{Hehai Lin$^{1}$\thanks{Equal contribution},
  Shilei Cao$^{2}$\footnotemark[1],
  Sudong Wang$^{1}$,
  Haotian Wu$^{1}$, \\
  \bf
  Minzhi Li$^{4,5}$,
  Linyi Yang$^{3}$, 
  Juepeng Zheng$^{2}$,
  Chengwei Qin$^{1}$\thanks{Corresponding to \href{mailto:chengweiqin@hkust-gz.edu.cn}{chengweiqin@hkust-gz.edu.cn}} \\
  $^{1}$The Hong Kong University of Science and Technology (Guangzhou), \\
  $^{2}$Sun Yat-sen University, $^{3}$Southern University of Science and Technology, \\
  $^{4}$National University of Singapore,
  $^{5}$Institute for Infocomm Research (I$^2$R), A*STAR
}

\begin{document}
\maketitle
\begin{abstract}
Existing multi-agent learning approaches explicitly foster collaboration among Large Language Models (LLMs) to build stronger multi-agent systems (MAS), yet they still rely on re-executing the MAS during inference. This contrasts with human cognition, wherein individuals can internalize insights from interactions to improve later independent reasoning. To investigate whether multi-agent interaction can enhance LLMs' independent problem-solving ability, we propose ILR (\textbf{I}nteractive \textbf{L}earning for LLM \textbf{R}easoning), a co-learning framework that integrates Dynamic Interaction and Perception Calibration. Dynamic Interaction adaptively selects cooperative or competitive strategies based on question difficulty and model capability, after which LLMs exchange information via Idea3 framework (Idea Sharing, Idea Analysis, and Idea Fusion), an interaction paradigm simulating human discussion, before producing final answers. Perception Calibration employs Group Relative Policy Optimization (GRPO) while integrating one LLM’s reward characteristics into another’s to strengthen interaction cohesion. We evaluate the effectiveness of ILR across three LLMs from two model families of varying scales on five mathematical, one coding, one general question answering, and one scientific reasoning benchmarks. 
We further investigate the advantage of Dynamic Interaction (i.e., boosting the robustness of stronger LLMs and surpassing pure strategy), and the scalability of ILR beyond two-model interactions. Code is available at \href{https://github.com/linhh29/Interactive-Learning-for-LLM-Reasoning}{https://github.com/linhh29/Interactive-Learning-for-LLM-Reasoning}.
\end{abstract}

\section{Introduction}

Efforts to enhance the reasoning capabilities of Large Language Models (LLMs) have largely relied on training paradigms such as Supervised Fine-Tuning (SFT)~\citep{achiam2023gpt}, Preference Learning (PL)~\citep{rafailov2023direct}, and Reinforcement Learning (RL)~\citep{schulman2017proximal, guo2025deepseek, zeng2025simplerl}. 
These methods allow LLMs to iteratively interact with data and refine their behavior, essentially engaging in trial-and-error learning to acquire problem-solving skills, which can be viewed as self-learning for LLMs. 
However, real-world knowledge acquisition is rarely an isolated activity~\citep{bloembergen2015evolutionary, canese2021multi}. Humans continuously exchange knowledge through collaborative learning, as in peer discussions within classroom settings. While \textit{single-agent learning} (self-learning) serves as the foundation of human education, \textit{multi-agent learning} represents a more advanced and often more effective paradigm: multiple learners bring diverse perspectives, challenge each other's reasoning, and provide mutual feedback, ultimately leading to deeper understanding and more robust solutions~\citep {kahveci2007interactive, hsiung2012effectiveness, zambrano2019effects, mende2021individual}. The same principle suggests that multi-agent learning can benefit LLMs\footnote{In our paper, ``Agent'' and ``LLM'' refer to the same entity}: by exposing models to diverse reasoning strategies and peer-based feedback, it may help them overcome individual blind spots and develop stronger problem-solving abilities. 

\begin{figure*}[t]
\begin{center}
\includegraphics[width=0.94\linewidth]{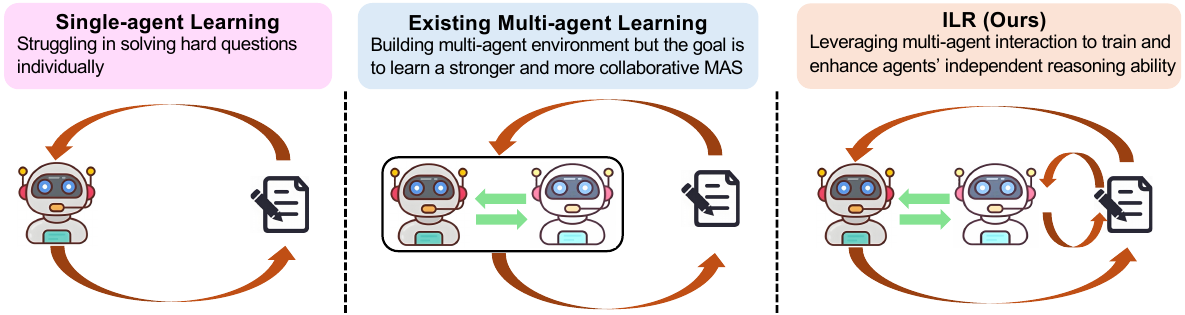}
\end{center}
\caption{Conceptual comparison of ILR, existing Multi-agent Learning, and Single-agent Learning. With ILR, LLMs engage in multi-agent learning at training time and can resolve questions independently at inference time.
}
\vspace{-10pt}
\label{fig: background}
\end{figure*}

Recent studies have explored multi-agent learning. For example, MALT~\citep{motwani2024malt} designs a sequential multi-agent system (MAS) consisting of Generator, Verifier, and Refiner, each independently trained to sample trajectories. ReMA~\citep{wan2025rema} introduces a hierarchical framework with a high-level agent responsible for problem decomposition and a low-level agent for concrete steps, trained alternately to achieve complementary expertise. MAPoRL~\citep{park2025maporl} proposes a Post-Co-Training framework to enhance collaboration alignment through debate. However, during inference, these methods are required to re-execute the MAS to obtain final solutions, a process misaligned with human cognition, where individuals improve reasoning through peer interactions and subsequently solve problems independently.


In this paper, we address this gap by treating each agent as an autonomous entity and investigating whether multi-agent learning can enhance an LLM's individual problem-solving capacity (see Figure~\ref{fig: background}). We propose \textbf{ILR} (\textbf{I}nteractive \textbf{L}earning for LLM \textbf{R}easoning), a co-learning framework consisting of two key components: \textit{Dynamic Interaction} and \textit{Perception Calibration}. 


The \textit{Dynamic Interaction} module simulates human discussion. For ``Dynamic'', when confronted with complex problems, humans tend to cooperate, whereas for simpler problems, they often compete to identify the most efficient solution~\citep{richard2002cooperation,schneider2011cooperation}. To emulate this behavior, an LLM estimates question difficulty through self-ranking and applies Item Response Theory~\citep{cai2016item, benedetto2023survey} to calculate the probability of solving it independently. If the probability is low, the model engages in cooperation; otherwise, it chooses competition. For ``Interaction'', we design a novel \textbf{Idea3} framework, comprising three sequential stages: Idea Sharing (each LLM proposes its own solution), Idea Analysis (each LLM analyzes and reflects on the peer's solution), and Idea Fusion (the insights are synthesized into a refined and potentially novel solution). Following Dynamic Interaction, the \textit{Perception Calibration} module is applied. Prior work~\citep{ma2024coevolving, park2025maporl} has shown that incorporating tailored reward signals can effectively guide LLMs toward better multi-agent learning. Instead of relying on predefined signals, we propose a fully automated mechanism that integrates one LLM's reward distribution characteristics, derived from answer group sampling on the same input, into another LLM's reward function. We then employ the standard GRPO~\citep{shao2024deepseekmath} to update each LLM based on calibrated rewards. 
This calibration allows LLMs to perceive the quality of peer-generated solutions and adapt their reasoning accordingly. We further validate the generalizability of ILR on another RL algorithm in Appendix~\ref{appendix: extendability}.

We evaluate the effectiveness of ILR across three LLMs from two model families of varying scales on five mathematical, one coding, one general question answering, and one scientific reasoning benchmarks. 
Experimental results demonstrate that ILR consistently outperforms both single-agent learning and multi-agent learning.
Our investigation further reveals three findings: (1) Dynamic Interaction enhances the robustness of stronger LLMs during multi-agent inference scenarios. Analyzing and reflecting on the peer's solutions reduces the probability of being misled by weaker LLMs when exchanging information.
(2) Dynamically determining interaction types can boost the efficacy of multi-agent learning and surpass pure cooperation or competition strategies.
(3) ILR has a positive scaling trend with a larger number of LLMs.
Our main contributions are summarized as follows:

\begin{itemize}
    \item Unlike prior work that focuses on improving system-level collaboration in multi-agent learning, we are the first to explore its potential for enhancing an individual LLM’s independent reasoning capability.
    \item Inspired by human interaction, we design a novel multi-agent learning framework ILR, which comprises Dynamic Interaction and Perception Calibration. 
    \item Experimental results validate ILR’s superiority over existing baselines. We further discover the advantage of Dynamic Interaction and the Scalability of ILR, which offer valuable insights toward future MAS design.
    
\end{itemize}


\section{Related Work}
\label{related_work}

Multi-agent learning~\citep{busoniu2006multi, han2024llm, li2024survey} first requires designing a multi-agent system that defines interaction paradigms among multiple agents, such as equi-level~\citep{chan2023chateval}, hierarchical~\citep{gronauer2022multi}, or nested structures~\citep{zhao2025sirius}. Then, within this architectural framework, distinct agents engage in interactive sampling to acquire experience, which subsequently undergoes optimization through learning algorithms. Therefore, we systematically review prior works from the following two perspectives: Multi-Agent Communication (interactive paradigms) and Multi-Agent Training (optimization methods).

\subsection{Multi-Agent Communication}
Traditionally, researchers employ recurrent neural networks (RNNs) as agents and utilize attention mechanisms to facilitate communication~\citep{yu2019review, ding2024learning, sun2024t2mac}. For instance, TarMAC leverages multi-head attention to enable agents to learn both message content and targeted recipient~\citep{das2019tarmac}. After the emergence of LLMs, researchers develop numerous explainable prompt-based multi-agent communication~\citep{lin2026unified}. Notable examples include Debate, where multiple agents articulate arguments culminating in a final answer through majority voting mechanisms~\citep{liang2023encouraging}, and Actor-Critic, where actor agents generate solutions subsequently evaluated by critic agents through iterative feedback processes~\citep{shinn2023reflexion, estornell2024acc, yuan2025reinforce}. 
However, existing communication collectively conceptualizes individual agents as components with optimization objectives centered on MAS performance. 

In contrast, our work treats each agent as an independent entity. 
We emulate human discussion dynamics through a novel Idea3 interaction, specifically designed to facilitate critical thinking communication among agents via its three-stage process: Idea Sharing, Idea Analysis, and Idea Fusion.

\subsection{Multi-Agent Training}
Conventional multi-agent training typically trains agents independently without awareness of other agents' states. Researchers employ multi-agent inference to collectively sample experiences, subsequently applying SFT or DPO to update individual agents independently, e.g., MALT~\citep{motwani2024malt}, Multiagent-FT~\citep{subramaniam2025multiagent}, and DEBATUNE~\citep{li2024can}. However, this static one-time sampling fundamentally compromises the dynamic nature of multi-agent interactions.
To address this limitation, recent advances in Multi-Agent Reinforcement Learning (MARL) have enabled continuous, real-time interaction sampling among agents~\citep{ma2024coevolving, chen2025improving, liao2025marft}. For example, 
MAPoRL~\citep{park2025maporl} implements a multi-agent proximal policy optimization algorithm, defining the agent state as the concatenation of interaction histories and incorporating manually predefined hyperparameters into rewards to incentivize collaboration.

Building on their design, we introduce a fully automated reward calibration,
enabling automatic peer perception without manual intervention.

\begin{figure*}[t]
\begin{center}
\includegraphics[width=1.0\linewidth]{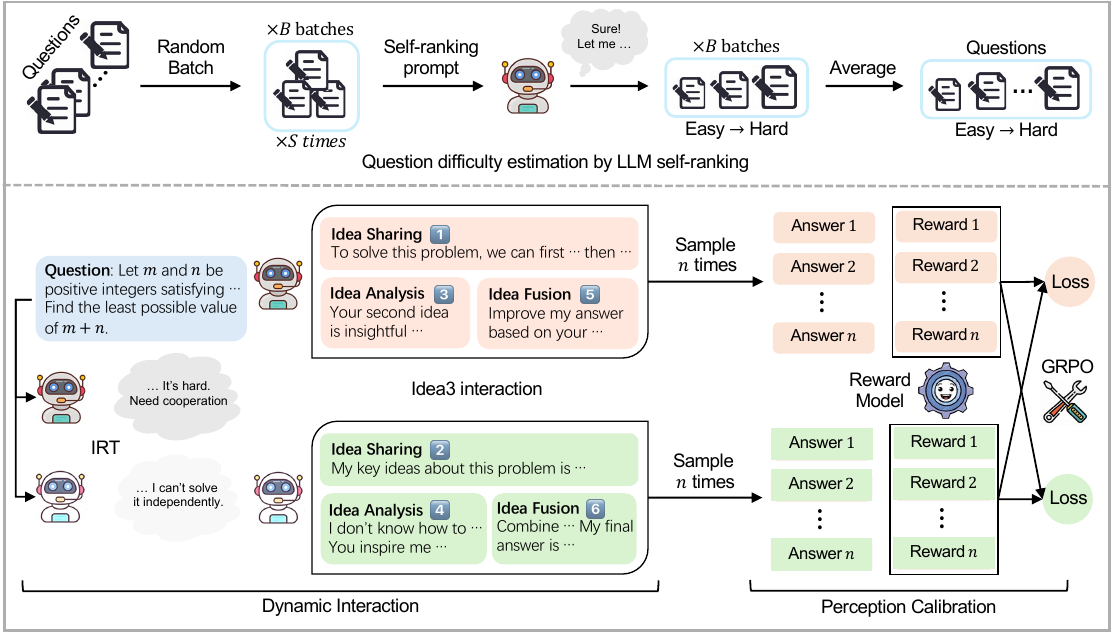}
\end{center}
\caption{Illustration of proposed ILR. \textbf{Top}:
LLMs first sort questions within each batch using the self-ranking prompt and then compute the average rank. \textbf{Bottom}: We depict the cooperation interaction here. If the question is too hard, LLMs will share their ideas, identify those complementary steps from other LLMs' ideas, and improve their answers. The competition interaction also follows Idea3, which only requires a minor change in prompts.}
\vspace{-10pt}
\label{fig:method}
\end{figure*}

\section{Methodology}


As illustrated in Figure~\ref{fig:method}, for each question, each LLM first dynamically selects an appropriate interaction mode and then engages in Idea3 interactions (Idea Sharing, Idea Analysis, and Idea Fusion) to produce a group of answers. Finally, a reward calibration allows LLMs to perceive the quality of peers' solutions, fostering more effective multi-agent learning. 
\revise{By doing so, we aim to more accurately simulate and study the human-like behaviors in existing LLMs within a context that mirrors real-world human learning scenarios. We provide the pseudocode of ILR in Appendix~\ref{appendix: pseudocode}.}

\subsection{Question Difficulty Estimation}

In real-world learning environments, such as classrooms, students often adapt their strategies to the complexity of the problem: for challenging tasks, they are more likely to collaborate, whereas for simpler ones, they tend to compete to demonstrate the efficiency and superiority of their solutions~\citep{richard2002cooperation, green2006children, schneider2011cooperation, fulop2022cooperation}. Both cooperation and competition can serve as drivers of multi-agent learning, which provides the underlying motivation for our Dynamic Interaction design. Since most problems lack explicit or continuous difficulty annotations, we adopt a self-ranking to estimate the question difficulty~\citep{wang2025make}.


Given a training dataset of $N$ questions, we divide it into $B$ random batches to avoid the long-context lost problem~\citep{liu2023lost}, with each batch containing $N' = \frac{N}{B}$ questions ($N' \ll N$). Using a self-ranking prompt (detailed in Appendix~\ref{appendix: prompt details}), the LLM is instructed to order the questions within each batch in ascending difficulty. The ranks are then normalized into difficulty scores, where the easiest question is assigned $\frac{1}{N'}$ (rank 1) and the hardest one is assigned $1.0$ (rank $N'$). Since a single random split provides only relative difficulty within its batch, we perform $S$ independent splits and average their results to obtain a more stable and robust difficulty estimate for the entire dataset. We set $N'$ and $S$ to 10 in our work.
For a set of LLMs $M=\{\mathcal{M}_{i} | i=0,1,...,m\}$ in a multi-agent learning scenario, the difficulty score $D_q$ of a question $q$ is computed as:
\begin{align}
D_{q}=\frac{1}{m}\sum_{i=1}^{m}\mathcal{M}_{i}(\frac{1}{S}\sum_{j=1}^{S}\frac{r_{q,j}}{N'})  
\end{align}
where $m$ is the number of LLMs, $\mathcal{M}_{i}(\cdot)$ denotes the estimation given by the $i$-th LLM, and $r_{q,j}$ is the rank of question $q$ in the $j$-th random split.

\subsection{Dynamic Interaction}
\label{sec: dynamic interaction}
When an LLM with reasoning ability $\gamma_{i}$
receives a question of difficulty level $D_{q}$, it can quantify the probability ($P_{q, i}$) of correctly answering the question using IRT~\citep{benedetto2023survey}:
\begin{align}
P_{q, i} = \frac{1}{1+e^{-1.7\times(\gamma_{i}-D_{q})}}
\end{align}
where the empirically derived coefficient 1.7 has been shown to yield reliable predictions across diverse conditions~\citep{baker2001basics, de2013theory, benedetto2023survey}. \revise{IRT is a mature framework from educational measurement designed to simultaneously model the relationship between three key variables: the ability of an individual, the difficulty of a question, and the probability of the individual correctly solving the question. This is highly analogous to our scenario, and previous works have investigated the intersection of IRT and AI/LLM~\citep{wang2023evaluating, lalor2024item, zhuang2023efficiently}, demonstrating its relevance and utility in our domain.}
Since $P_{q, i}=0.5$ when $\gamma_{i}=D_{q}$, we adopt 0.5 as the decision boundary between different interaction modes. We average the probability of $m$ LLMs to derive the overall probability ($P_{q}$) of independently solving question $q$ and determine the interaction mode.
We study the effect of mixed-type interaction for each question in Appendix~\ref{appendix: mixed types interaction}.
\begin{align}
\text{Mode}&= \begin{cases}
\text{Cooperation} & \text{if} \: P_{q} < 0.5\\
\text{Competition} & \text{if} \: P_{q} \ge  0.5
\end{cases}
\end{align}

To simulate human discussion, we design a novel and unified three-stage Idea3 interaction for multi-agent communication: Idea Sharing (each LLM proposes its own solution), Idea Analysis (each LLM analyzes and reflects on the peer's solution), and Idea Fusion (the insights are synthesized into a refined and potentially novel solution). For different modes, we only need to slightly modify the prompt to inject the corresponding signal (see Appendix~\ref{appendix: prompt details}). \revise{Unlike debate frameworks, which directly encourage one LLM to take another's output as advice without further thinking, our Idea3 is designed to foster critical thought during communication. We show two cases in Appendix~\ref{appendix: case study}.}

\textbf{Idea Sharing.} Each LLM begins by presenting its problem-solving strategy, explaining the reasoning process and methods employed to address the given problem. For example, when solving a complex algebraic equation, one model might focus on factoring, while another may rely on graphical analysis. This stage produces the \textit{initial answer}.

\textbf{Idea Analysis.} Subsequently, LLMs engage in a critical evaluation of each other's proposed methods. In the cooperation mode, they may identify complementary strengths from different approaches, such as combining graphical insights with algebraic manipulation to generate a more comprehensive solution. In the competition mode, however, they rigorously assess the merits and limitations of the shared strategies. For example, one LLM might argue that the factoring approach, while effective, overlooks potential solutions that could be derived from the quadratic formula, thereby revealing a potential improvement.

\textbf{Idea Fusion.} Finally, LLMs synthesize the insights gained during previous analyses to generate a refined answer. This may involve integrating the most effective elements of both approaches, yielding a solution that not only accurately addresses the problem but also leverages complementary techniques from each LLM. For example, the final resolution to the algebraic equation might incorporate both the graphical representation for visual clarity and the algebraic methods for precision, culminating in a solution that is both robust and comprehensible. 
This stage's output is the \textit{updated answer}.

Prior research has noted that inter-agent communication may introduce noise into 
final outputs~\citep{pan2025multiagent, zhang2025agent}. To address this, we adopt a label-based selection mechanism: the initial answer is retained only if it is correct and the updated answer is incorrect; in all other cases, the updated answer is chosen. 
\revise{
Finally, the rollout experience only contains this chosen solution, and the reward is calculated solely based on this ``final answer''. This design choice can ensure: (1) To avoid a potential prompt shift. Since the rollout only contains a higher-quality step-by-step solution without intermediate results, the LLM is trained on the same type of data that it is expected to produce during inference. Therefore, there is no misalignment between the training and inference prompts. (2) To ensure a fair comparison with single-agent learning baselines. Including the rich information from the interaction trajectory in the reward signal could introduce a confounding variable, making it difficult to determine whether performance improvements stem from ILR or simply from the additional information in the trajectory. 
}

\subsection{Perception Calibration}
\label{Perception Calibration}
Prior works rely on adding manually predefined signals to rewards~\citep{park2025maporl}, which are discrete and coarse-grained, and in turn limit the scalability of reward shaping. In contrast, we introduce a fully automated method that integrates the distributional characteristics of one LLM's reward data into another LLM's reward function. 

For a given input question, $m$ LLMs each perform $n$ sampling rounds, producing $m$ groups of responses, with $n$ answers in each group. A reward model is first used to assign initial rewards $R$ to all responses. Each group is then summarized by its maximum ($R_{max}$), minimum ($R_{min}$), and average ($R_{avg}$) scores, collectively reflecting the model's overall answer quality for that question. These statistics are arithmetically normalized and injected into the reward shaping process of peer models, yielding the final reward $\bar{R}$. For example, the $k$-th final reward of LLM $i$ is computed as:
\begin{align}
    \bar{R}_{i,k} = R_{i,k} + \sum_{l \in M \setminus \{\mathcal{M}_{i}\} }\text{clip}(\delta_{i,l}, -U, U)
\end{align}

where $\delta_{i,l} = \frac{R_{i,k} - R_{l, avg}}{R_{l,max} - R_{l, min}}$, $U = \frac{1}{m-1}$, $\text{clip}(\cdot)$ is a stabilization operation to prevent extreme values. We further discuss the suitability of $\text{clip}(\cdot)$ in Appendix~\ref{clip ratio} and study the effect of z-score variant of $\delta_{i,l}$ in Appendix~\ref{appendix: z-score}. We then apply standard GRPO to optimize LLMs using calibrated rewards.



\section{Experimental Setting}
\label{sec: experimental setting}
\textbf{Test Models.}
We conduct experiments on three representative LLMs spanning two series and two scales: Llama-3.1-8B-Instruct~\citep{dubey2024llama}, Qwen2.5-7B-Instruct, and Qwen2.5-14B-Instruct~\citep{team2024qwen2}. We organize these models into three groups and apply ILR within each. We further study a homogeneous setting (same LLM with different input prompts) in Appendix~\ref{appendix: same llm different prompts}.
\begin{itemize}
    \item Group1 (different \textbf{series}): Llama-3.1-8B-Instruct and Qwen2.5-7B-Instruct
    \item Group2 (different \textbf{series} \& \textbf{scale}): Llama-3.1-8B-Instruct and Qwen2.5-14B-Instruct
    \item Group3 (distinct \textbf{scale}): Qwen2.5-7B-Instruct and Qwen2.5-14B-Instruct
\end{itemize}

\textbf{Benchmarks.}
For ILR \textit{training}, we use the MATH~\citep{hendrycks2021measuring} following~\citep{zeng2025simplerl}
and randomly select 1,000 samples as the validation set to estimate each LLM's $\gamma_{i}$ offline before training, while the other 11,000 samples are used for training. We compare this one-time measurement strategy ($\gamma_{i}$) with a periodic-update strategy in Appendix~\ref{appendix: periodic}.
For ILR \textbf{evaluation}, we conduct a comprehensive assessment across multiple mathematical reasoning benchmarks, which encompass both standard benchmarks, including GSM8K~\citep{cobbe2021training}, MATH-500~\citep{hendrycks2021measuring}, Minerva Math~\citep{lewkowycz2022solving}, and Olympiad Bench~\citep{he2024olympiadbench}, and a competition-level benchmark AIME24\&25~\citep{aime}, consisting of AIME2024 and AIME 2025. To further assess generalization beyond math, we evaluate ILR and baselines on a code generation benchmark MBPP~\citep{austin2021program}, \camera{a general QA benchmark NQ~\citep{kwiatkowski2019natural}, and a scientific reasoning benchmark MMLU~\citep{hendrycks2020measuring}}. \revise{Note that except for Table 1, which is averaged across 5 seeds, all the other results are based on seed 0.}

\begin{table*}[t!]
\centering
\small
\begin{tabular}{@{}l|cccccc@{}}
\toprule
 & GSM8K & MATH-500 & \begin{tabular}[c]{@{}c@{}}Minerva\\ Math\end{tabular}& \begin{tabular}[c]{@{}c@{}}Olympiad\\ Bench\end{tabular} & \begin{tabular}[c]{@{}c@{}}AIME\\ 24\&25\end{tabular}  & Avg \\ \midrule
Llama-3.1-8B-Instruct &82.38&48.60&23.31&13.30&2.33&33.99\\
SFT &85.17&49.68&23.90&16.68&3.00&35.69\\
DPO &86.06&50.12&24.63&17.36&2.67&36.17\\ 
PPO &86.17&52.44&26.03&19.26&4.00&37.58\\
GRPO &85.70&52.32&26.69&19.79&4.00&37.70\\
GSPO &85.82&53.76&27.43&18.90&5.00&38.18\\ 
Reinforce++ &86.88&51.40&29.78&18.52&3.33&37.98\\ 
DebateFT-Group1 &84.91&49.60&25.14&17.93&2.67&36.05\\
DebateFT-Group2 &84.82&49.24&24.19&16.89&2.67&35.56\\ 
\camera{MAPoRL-Group1} & 86.81 & 53.00 & 30.88 & 21.93 & 6.67 & 39.86 \\
\camera{MAPoRL-Group2} & 87.41 & 53.60 & 31.25 & 21.48 & 5.00 & 39.75 \\
\camera{COPY-Group1} & 86.88 & 54.40 & 28.68 & 19.70 & 5.00 & 38.93 \\
\camera{COPY-Group2} & 86.58 & 53.00 & 29.41 & 19.11 & 6.67 & 38.95 \\
\rowcolor{lightgreen}
ILR-Group1 &\textbf{88.40}&\textbf{54.56}&30.37&\textbf{22.07}&\textbf{9.00}&\textbf{40.88}\\
\rowcolor{lightgreen}
ILR-Group2 &87.38&54.36&\textbf{32.28}&21.51&6.67&40.44\\ 
\midrule

Qwen2.5-7B-Instruct &91.83&74.84&40.95&36.83&9.33&50.76\\
SFT &91.90&75.92&41.25&37.24&11.33&51.53\\
DPO &92.11&75.72&42.43&37.54&10.33&51.63\\
PPO &92.12&76.44&43.31&38.07&11.67&52.32\\
GRPO&92.48&75.88&41.91&38.19&13.00&52.29\\
GSPO
&92.33&76.80&42.06&38.34&13.66&52.64\\ 
Reinforce++ &92.57&76.56&42.13&38.22&12.67&52.43\\
DebateFT-Group1 &92.16&75.96&41.25&37.28&10.33&51.40\\
DebateFT-Group3 &92.17&76.44&41.11&37.36&12.33&51.88\\ 
\camera{MAPoRL-Group1} & 92.03 & 76.60 & 44.12 & 38.67 & 15.00 & 53.28 \\
\camera{MAPoRL-Group3} & 92.10 & 77.20 & 42.65 & 39.41 & 13.33 & 52.94 \\
\camera{COPY-Group1} & 92.57 & 77.00 & 42.65 & 39.11 & 16.33 & 53.53 \\
\camera{COPY-Group3} & 92.27 & 77.20 & 43.01 & 38.52 & 15.00 & 53.20 \\
\rowcolor{lightgreen}
ILR-Group1 &\textbf{93.00}&77.00&44.56&\textbf{39.50}&16.33&54.08\\
\rowcolor{lightgreen}
ILR-Group3 &92.88&\textbf{77.80}&\textbf{44.73}&39.17&\textbf{17.00}&\textbf{54.31}\\
\midrule

Qwen2.5-14B-Instruct &94.83&80.08&46.03&40.21&12.33&54.70\\
SFT &94.98&80.20&46.32&40.86&13.00&55.07\\
DPO &95.09&80.84&47.28&40.59&14.33&55.63\\
PPO &95.06&80.88&48.45&41.10&16.67&56.43\\
GRPO &94.86&80.48&47.43&42.11&16.00&56.17\\
GSPO &95.30&80.12&48.82&42.96&16.34&56.71\\ 
Reinforce++ &95.03&80.68&48.31&41.39&16.67&56.41\\
DebateFT-Group2 &94.85&80.56&46.62&41.15&13.67&55.37\\
DebateFT-Group3 &94.92&80.76&47.57&41.48&12.00&55.35\\ 
\camera{MAPoRL-Group2} & 95.22 & 80.80 & 49.26 & 42.41 & \textbf{20.00} & 57.54 \\
\camera{MAPoRL-Group3} & 95.45 & 81.40 & 50.00 & 42.67 & 18.33 & 57.57 \\
\camera{COPY-Group2} & 95.08 & 81.20 & 50.00 & 42.81 & 16.67 & 57.15 \\
\camera{COPY-Group3} & 95.28 & 81.80 & 48.90 & 43.11 & 18.33 & 57.48 \\
\rowcolor{lightgreen}
ILR-Group2 &\textbf{95.50}&81.56&49.85&43.11&\textbf{20.00}&58.00\\
\rowcolor{lightgreen}
ILR-Group3 &95.42&\textbf{82.32}&\textbf{50.07}&\textbf{43.53}&19.66&\textbf{58.20} \\ 
\bottomrule
\end{tabular}%
\caption{The quantification comparison (accuracy \%) of ILR and other baselines.}
\label{tab:mian_result}
\end{table*}

\textbf{Baselines.}
We compare our ILR with six single-agent learning (self-learning) baselines, including SFT~\citep{achiam2023gpt}, DPO~\citep{rafailov2023direct}, PPO~\citep{schulman2017proximal}, GRPO~\citep{shao2024deepseekmath}, GSPO~\citep{zheng2025group}, and Reinforce++~\citep{hu2025reinforce++}, and three multi-agent learning baselines, including DebateFT~\citep{subramaniam2025multiagent}, \camera{COPY~\citep{ma2024coevolving}, and MAPoRL~\citep{park2025maporl}}. These baselines represent widely adopted and empirically effective approaches.
Details about baselines, hyperparameters, training, and evaluation are provided in Appendix~\ref{appendix: Implementation Details} and Appendix~\ref{appendix: statistics of benchmarks}. We further compare ILR with more self-learning baselines like self-reflection learning and advanced GRPO variants in Appendix~\ref{appendix: more self-learning methods}.

\section{Results and Analysis}
This section presents our main findings on ILR. Additional discussions on role assignment and application scenarios are provided in Appendix~\ref{appendix: discussion}. Further analyses, including z-score reward normalization, periodic update of $\gamma_i$, mixed-type interactions, cost analysis, homogeneous LLM settings, and another RL algorithm, are reported in Appendix~\ref{appendix:more ablation study}.


\subsection{LLMs Perform Better Through ILR}
\label{sec: llms perform better through ILR learning}

\begin{table*}[]
\centering
\begin{tabular}{@{}ll|cccccc@{}}
\toprule
Benchmark & Model & Base & DPO  & GRPO & ILR-G1 & ILR-G2  & ILR-G3 \\ \midrule
\multirow{3}{*}{MBPP} & Llama-3.1-8B-Instruct & 54.00 & 56.40 & 56.80 & 57.40 & \textbf{57.60} & - \\ 
 & Qwen2.5-7B-Instruct & 64.80 & 65.20 & 65.20 & 65.60 & - & \textbf{66.20} \\ 
 & Qwen2.5-14B-Instruct & 71.40 & 71.60 & 71.80 & - & \textbf{72.40} & 71.60 \\ \midrule
\multirow{3}{*}{\camera{NQ}} & Llama-3.1-8B-Instruct & 51.12 & 52.32 & 52.11 & 54.77 & \textbf{55.31} & - \\ 
 & Qwen2.5-7B-Instruct & 40.23 & 41.38 & 41.80 & 43.23 & - & \textbf{43.60} \\ 
 & Qwen2.5-14B-Instruct & 49.13 & 49.60 & 50.00 & - & 52.45 & \textbf{53.87} \\ \midrule
\multirow{3}{*}{\camera{MMLU}} & Llama-3.1-8B-Instruct & 67.11 & 67.55 & 68.77 & \textbf{70.65} & 70.12 & - \\ 
 & Qwen2.5-7B-Instruct & 52.15 & 52.54 & 52.56 & 54.00 & - & \textbf{54.14} \\ 
 & Qwen2.5-14B-Instruct & 78.33 & 78.45 & 78.60 & - & 79.66 & \textbf{80.22} \\ \bottomrule
\end{tabular}%
\caption{Out-of-domain evaluation of ILR, DPO, and GRPO on MBPP (Pass@1), NQ (Accuracy), and MMLU (Accuracy). G$i$ means Group$i$. Compared with representative baselines, ILR further improves the performance.}
\label{tab:ood_mbpp}
\end{table*}

\begin{table*}[]
\centering
\begin{tabular}{@{}l|cc|cc|cc@{}}
\toprule
& \multicolumn{2}{c|}{Llama-3.1-8B-Instruct} & \multicolumn{2}{c|}{Qwen2.5-7B-Instruct} & \multicolumn{2}{c}{Qwen2.5-14B-Instruct} \\
 &Group1 & Group2 & Group1 & Group3 & Group2 & Group3 \\ \midrule
ILR &\textbf{41.51}&\textbf{41.10}&\textbf{54.44}&\textbf{54.59}&\textbf{58.95}&\textbf{59.30}\\
DI-only &39.12&39.25&53.95&53.23&58.04&58.57\\
PC-only &40.14&38.41&54.04&53.91&57.66&58.07\\
\bottomrule
\end{tabular}%
%
\caption{Ablation Study of ILR. We report the average accuracy (\%) of five mathematical evaluation benchmarks. DI, PC represent Dynamic Interaction and Perception Calibration.}
\label{tab: ablation study}
\end{table*}

\textbf{Overall Comparison.} Table~\ref{tab:mian_result} presents the quantitative comparison on mathematical benchmarks and Table~\ref{tab:ood_mbpp} depicts the out-of-domain performance on MBPP, NQ, and MMLU. The results show that ILR's multi-agent learning framework consistently enhances the independent problem-solving capability of individual LLMs. 
From Table~\ref{tab:mian_result}, we highlight two further insights: (1) \textbf{ILR Promotes Complex Reasoning.} On the competition-level dataset AIME24\&25 (comprising AIME24 and AIME25), ILR significantly improves LLMs' capability in solving complex problems, which demonstrates that both weaker and stronger models can benefit from ILR's multi-agent learning framework, enabling them to independently tackle challenging reasoning tasks.
(2) \textbf{Balanced Grouping Improves Learning.} Each LLM is trained with ILR under two different grouping settings.
Across comparisons, models achieve stronger results when paired with peers of more similar initial reasoning ability. For example, Llama-3.1-8B-Instruct performs better in ILR-Group1 with Qwen2.5-7B-Instruct (40.88\% vs. 40.44\% in Group2). We attribute this phenomenon to the fact that excessive initial performance disparities may lead to imbalanced interactions where the stronger LLM overwhelmingly dominates the entire process.
While this pattern consistently emerges across three groups, the observed performance differences are modest. We leave more comprehensive empirical validation of this finding to future research.

\textbf{Ablation Study.} To further analyze contributions of each component, we conduct an ablation study and report the average accuracy in Table~\ref{tab: ablation study}. Removing either component leads to consistent performance drops across all models, underscoring their joint contribution to the overall performance.


\subsection{Dynamic Interaction Enhances Stronger LLMs' Robustness}
In Section~\ref{sec: llms perform better through ILR learning}, we demonstrate how multi-agent learning through ILR training strengthens the independent reasoning abilities of individual LLMs. Here, we evaluate the effectiveness of Idea3 communication during the inference stage. In pure inference scenarios where ground-truth labels are unavailable, we employ a summarization prompt to synthesize the initial and updated responses, thereby mitigating noise from multi-agent interactions. Specifically, for a given input question, two LLMs first engage in Idea3 communication, after which each model evaluates both its own initial answer and the updated answer to produce a final prediction. For comparison, we also include Debate, which treats other agents' outputs as additional advice to inform final answer generation. Full prompt details are provided in Appendix~\ref{appendix: prompt details}.


Table~\ref{tab:multi-agent_inference} shows the multi-agent inference result on MATH-500 \revise{using untrained LLMs}. We consistently observe that within different groups, Debate is more beneficial for weaker LLMs, while our Idea3 enhances the robustness of stronger LLMs by making them less susceptible to low-quality responses generated from weaker LLMs during multi-agent communication. We attribute this phenomenon to two primary reasons: \textbf{First}, for weaker LLMs, Debate directly incorporates stronger models' answers as additional guidance, enabling them to refine their outputs, which often results in a more significant improvement. \textbf{Second}, for stronger LLMs, Debate similarly compels them to consider answers from weaker LLMs, which are usually lower in quality and can potentially degrade performance. In contrast, our Idea3 prompts stronger LLMs to critically evaluate and selectively integrate peer contributions, filtering out noise and thereby improving robustness.
\revise{
We further conduct multi-agent inference using ILR-trained LLMs as base models. Results show that ILR-trained LLMs achieve better performance than untrained LLMs in the same multi-agent inference setup, which indicates that ILR does not diminish an LLM's ability to collaborate effectively at deployment.
}

\subsection{Cooperation vs. Competition}

In ILR training, we employ Item Response Theory (IRT) to dynamically determine interaction types, i.e., cooperation or competition. To further investigate the influence of cooperation, we vary the cooperation ratio ($p$) from 0.0 to 1.0 in increments of 0.2. Here, $p=0.0$ corresponds to full competition, $p=1.0$ to full cooperation, and intermediate values designate the first $p$-proportion of questions (ranked by difficulty) as cooperative, with the remainder treated competitively. Due to training costs, we restrict this study to Group1.

\begin{table*}[t!]
\centering
\begin{tabular}{@{}l|cc|cc|cc@{}}
\toprule
\multirow{2}{*}{\begin{tabular}[c]{@{}c@{}}Inference\\ Paradigm\end{tabular}}& \multicolumn{2}{c|}{Llama-3.1-8B-Instruct} & \multicolumn{2}{c|}{Qwen2.5-7B-Instruct} & \multicolumn{2}{c}{Qwen2.5-14B-Instruct} \\
 &Group1 & Group2 & Group1* & Group3 & Group2* & Group3* \\ \midrule
Single &49.80&49.80&75.60&75.60&81.20&81.20\\
Debate &\textbf{64.00}&\textbf{66.20}&74.60&\textbf{80.00}&79.20&81.00\\
Idea3 (Ours) &63.40&62.00&\textbf{75.60}&77.80&\textbf{79.80}&\textbf{82.00}\\ \midrule

\revise{ILR-Single} &\revise{55.80}&\revise{55.20}&\revise{77.60}&\revise{78.00}&\revise{81.80}&\revise{82.60}\\
\revise{ILR-Debate} &\revise{\textbf{65.60}}&\revise{\textbf{71.20}}&\revise{81.20}&\revise{\textbf{81.00}}&\revise{79.40}&\revise{81.20}\\
\revise{ILR-Idea3 (Ours)} &\revise{65.00}&\revise{67.80}&\revise{\textbf{76.00}}&\revise{78.80}&\revise{\textbf{80.40}}&\revise{\textbf{82.40}} \\

\bottomrule
\end{tabular}%
\caption{Multi-agent inference results on MATH-500. `Single' denotes the single-agent inference performance of the base models, and * indicates the stronger LLM within each group. For Debate and Idea3, the better-performing result is highlighted in bold. \revise{ILR- means we utilize LLMs after ILR-training as base models.}}
\label{tab:multi-agent_inference}
\end{table*}

\begin{figure}[t]
\includegraphics[width=0.48\textwidth]{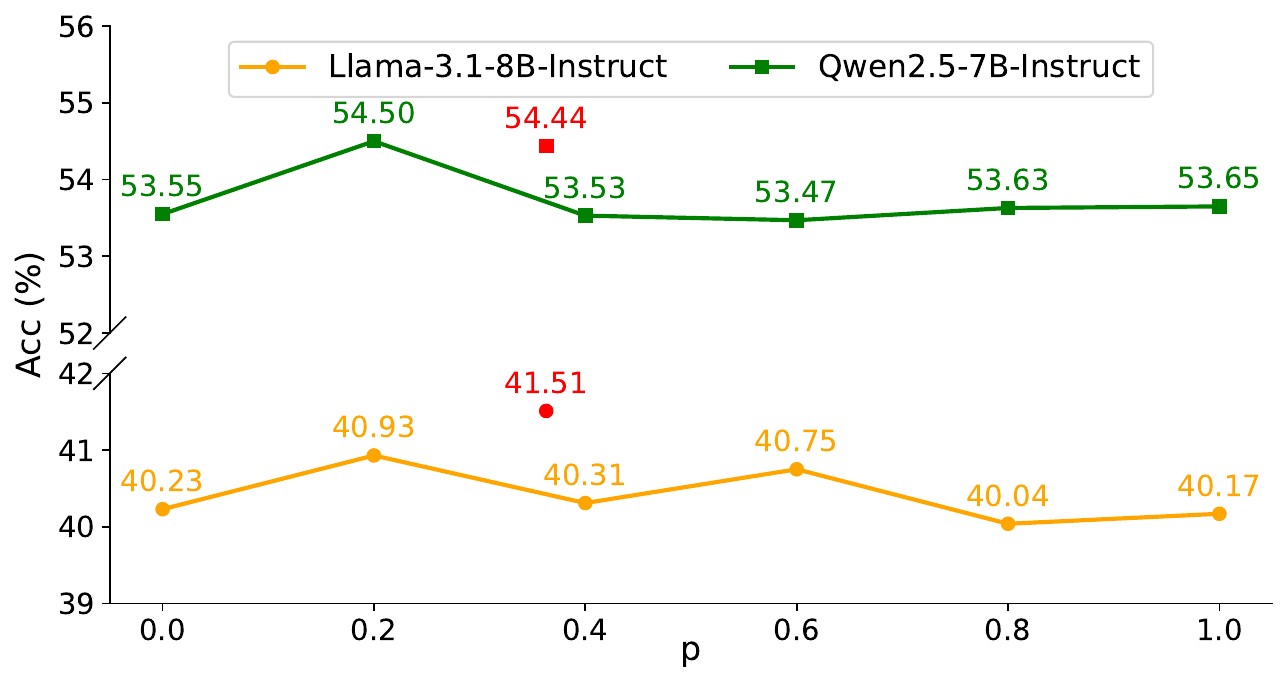}
\caption{Average accuracy of Group1 under varying cooperation ratios. IRT is marked in red.}
\vspace{-12pt}
\label{fig: different proportion of cooperation}
\end{figure}

Figure~\ref{fig: different proportion of cooperation} shows the results, with IRT highlighted in red. Two key findings emerge for dynamic interaction design:
(1) \textbf{Suboptimality of Extreme Strategies.} Relying solely on competition or cooperation is suboptimal for ILR, underscoring the necessity of adaptive interaction in multi-agent learning. This is intuitive: for challenging problems, cooperation allows LLMs to leverage complementary strengths and produce more comprehensive solutions, whereas for simpler tasks that can be effectively solved independently, excessive cooperation provides little benefit and may even introduce noise into the final outputs.
(2) \textbf{Configuration of $p$.} The optimal cooperation ratio $p$ requires careful design. One option is to manually partition data into subsets and tune $p$, but this is costly. IRT offers a practical alternative by approximating problem difficulty and aligning it with model reasoning capability. Although not always optimal (e.g., for Qwen2.5-7B-Instruct), it achieves competitive results while eliminating manual intervention. This demonstrates the feasibility of IRT as a principled mechanism for integrating problem difficulty with LLM reasoning abilities. Future work may enhance robustness by incorporating additional conditional parameters into the IRT formulation.

\revise{
\subsection{Scalability of ILR}

On balance of training costs and effectiveness, we primarily conduct experiments on three pairwise groups. To further evaluate the scalability of ILR beyond two-model interactions, we introduce a tri-model configuration, designated as Group 4 (ILR-G4). This experiment group integrates Llama-3.1-8B-Instruct, Qwen2.5-7B-Instruct, and Qwen2.5-14B-Instruct into a unified collaborative framework. The primary objective is to investigate the scalability of ILR with more than two LLMs.

As evidenced in Table~\ref{tab:three_llms}, the deployment of three LLMs results in a consistent and monotonic performance improvement across all constituent models compared to their dyadic counterparts (ILR-G1, G2, and G3). All LLMs surpass their best two-model performance by nearly 1.0\%, which suggests that ILR effectively leverages the heterogeneity of larger model groups. The addition of a third model does not introduce noise or redundancy; rather, it provides complementary reasoning paths that further refine the consensus, confirming the robustness and positive scaling trend of our ILR. This presents an intriguing finding. Future work could focus on mitigating the training costs associated with multi-agent reinforcement learning, thereby enabling the investigation of interactive learning behaviors across broader ensembles of LLMs.

}

\begin{table}[t]
\centering
\setlength{\tabcolsep}{3.5pt}
\begin{tabular}{@{}l|cccc@{}}
\toprule
Model  & ILR-G1 & ILR-G2  & ILR-G3 & ILR-G4 \\ \midrule
Llama &41.51&41.10&-&\textbf{42.48}\\ 
Qwen7B &54.44&-&54.59&\textbf{55.12}\\ 
Qwen14B &-&58.95&59.30&\textbf{59.95}\\ \bottomrule
\end{tabular}%
\caption{\revise{Scaling trend of ILR (two vs. three LLMs per group). G$i$ denotes Group$i$. We report the average accuracy on five mathematical evaluation benchmarks.}}
\label{tab:three_llms}
\vspace{-10pt}
\end{table}

\section{Conclusion}

In this paper, we investigate whether interactive learning among multiple LLMs (\textit{multi-agent learning}) can outperform traditional self-learning (\textit{single-agent learning}). 
Inspired by real-world human interaction, we propose ILR, a novel framework built on two key components: Dynamic Interaction and Perception Calibration. 
Extensive experiments across different model series and scales demonstrate the effectiveness of ILR, showing that interactive learning consistently yields greater performance improvements than self-learning. We further investigate the advantage of Dynamic Interaction and the Scalability of ILR beyond two-model interactions.
These findings align with human learning patterns and provide insights into analyzing high-level, human-like behaviors in LLMs.

\section*{Limitations}
Our current reliance on intrinsic difficulty may be less designable for creative generation tasks, as even simple questions in this domain benefit significantly from the stylistic diversity of cooperation strategy. Therefore, additional conditioning variables are required to extend the robustness of Dynamic Interaction and the scenario of ILR.

\bibliography{custom}

\appendix

\section{Motivation of ILR}
\camera{
The rationale of the proposed ILR is intricately designed to function as a cohesive pipeline that mirrors humans' interactive learning. In real-world peer discussions, individuals sequentially experience three cognitive stages: first, upon encountering a problem, they instinctively assess its complexity (Difficulty Estimation). Second, based on this assessment, they determine their interaction strategy, competing to find the most elegant solution for simple tasks, or cooperating to overcome challenging ones (Dynamic Interaction with the Idea3 framework). Finally, after the discussion, individuals reflect on and are incentivized by the quality of their peers' solutions (Perception Calibration). 
}

\section{Experiment Details}

\subsection{Prompt Details}
\label{appendix: prompt details}

\begin{tcolorbox}[
  enhanced,
  breakable,
  fonttitle = \small\bfseries, 
  title=Self-ranking Prompt for Difficulty Estimation,
  colframe=Salmon!90!Black,   
  colback=Salmon!20,     
  boxrule=1pt,
  boxsep=2pt,
  left=5pt,
  right=5pt,
  fontupper=\footnotesize,
  halign title = flush center
]
\textbf{Ranking Prompt1:}

Your task is to rank the given questions from easy to hard based on their difficulty level. Questions to be evaluated: \{$Q1, Q2, ..., QN'$\}.

\textbf{Ranking Prompt2:}

You will be given a batch of questions. Your task is to rank them from easy to hard based on their difficulty level. You should carefully horizontally compare the given questions in order to assign a suitable ranking place to each question. Please make sure you read and understand these instructions carefully. Please keep this document open while reviewing, and refer to it as needed. Questions to be evaluated: \{$Q1, Q2, ..., QN'$\}.

\textbf{Ranking Prompt3:}

You need to analyze and rank questions $Q1$-$QN'$ by their difficulty level based on these criteria: (1) Cognitive load required. If a problem involves multiple steps, it will have a higher cognitive load than a problem with a single step. (2) Knowledge depth needed. Problems involving the deduction of complex formulas require deeper professional knowledge. (3) Typical error rates. For example, problems that tend to overlook a certain prerequisite or calculation step will have a relatively high error rate. Please first score each question (1-10 scale) on three dimensions above, then calculate the average score, and rank these questions by final scores. Questions to be evaluated: \{$Q1, Q2, ..., QN'$\}.

\textbf{Format Prompt:}

After analyzing all the questions, please give all the ranking places (from easy to hard) in order, following the template ``Ranking: $[Q$\{number of the easiest question\},..., $Q$\{number of the hardest question\} $]$''.
\end{tcolorbox}

\begin{tcolorbox}[
  enhanced,
  breakable,
  fonttitle = \small\bfseries, 
  title=Prompt for Idea3 communication,
  colframe=Salmon!90!Black,   
  colback=Salmon!20,     
  boxrule=1pt,
  boxsep=2pt,
  left=5pt,
  right=5pt,
  fontupper=\footnotesize,
  halign title = flush center
]
\textbf{Idea Sharing (Cooperation\&Competition):}

Question: \{Input question\}

Please reason step by step, and your final answer should be in the form boxed\{answer\} given at the end of your response.

\textbf{Idea Analysis (Cooperation):}

Partner's Contribution: \{Ideas from other LLMs\}

Collaboratively analyze the key steps in the partner's contribution, identify those steps that can help you improve your answer, and serve as additional advice.

\textbf{Idea Analysis (Competition):}

Opponent's Solution: \{Ideas from other LLMs\}

Critically analyze the opponent's ideas, identify the weaknesses and strengths of his ideas.

\textbf{Idea Fusion (Cooperation\&Competition):}

Based on the above analysis, give an updated answer to the Original Question: \{Input question\}.  Please reason step by step, and your final answer should be in the form boxed\{answer\} given at the end of your response.

\end{tcolorbox}

\begin{tcolorbox}[
  enhanced,
  breakable,
  fonttitle = \small\bfseries, 
  title=Prompt for Evaluation,
  colframe=Salmon!90!Black,   
  colback=Salmon!20,     
  boxrule=1pt,
  boxsep=2pt,
  left=5pt,
  right=5pt,
  fontupper=\footnotesize,
  halign title = flush center
]
\textbf{Single-agent Evaluation:}

Please reason step by step, and your final answer should be in the form boxed\{answer\} given at the end of your response.

\textbf{Multi-agent Evaluation:}

\textbf{Debate:}

Here are solutions from other agents:

One agent response: \{other\_agent\_response\}

Using each response as additional advice based on the correctness of each response. Can you give an updated bullet-by-bullet answer to \{Input question\}. Please reason step by step, and your final answer should be in the form boxed\{answer\} given at the end of your response.

\textbf{Summarization (After Idea3 communication):}

The original question is \{Input question\}. There are two solutions you provided:

Solution 1: \{Initial answer\}

Solution 2: \{Updated answer\}

Please answer the original question step-by-step based on these two solutions, and your final answer should be in the form boxed\{answer\} given at the end of your response.
\end{tcolorbox}

To mitigate potential prompt bias, we utilize three different self-ranking prompts of varying levels of granularity for each question and average the rankings to obtain the final estimation.

\camera{
Furthermore, we add a post-validation (see Table~\ref{tab:correlation}) to study the effectiveness of our difficulty estimation, which measures the correlation of self-ranking difficulty and empirical ``true solve rate'' derived from 20 samplings per question on the validation set (1,000 samples). The resulting high correlation coefficient between the self-ranking scores and the actual success rates confirms that our estimation mechanism accurately and robustly reflects the true difficulty relative to the LLM's capabilities. Specifically, the first three columns report the correlation between each individual llm's self-ranking scores and its respective solve probabilities, while the final ``Avg'' column presents the correlation derived from the averaged self-ranking scores and averaged solve probabilities across all llms.

\begin{table*}[h]
\centering
\begin{tabular}{@{}cccc@{}}
\toprule
Meta-Llama-3.1-8B-Instruct & Qwen2.5-7B-Instruct & Qwen2.5-14B-Instruct & Avg \\ \midrule
0.7138&0.6905&0.7254&0.7522\\ 
\bottomrule
\end{tabular}%
\caption{The Pearson correlation of self-ranking difficulty and empirical ``true solve rate''.}
\label{tab:correlation}
\end{table*}

}

\subsection{Implementation Details}
\label{appendix: Implementation Details}
\textbf{Baselines:} For multi-agent learning, direct comparisons are challenging because existing methods typically train specialized LLMs with complementary roles for problem-solving. To enable fair evaluation, we introduce a minor modification to Multiagent-FT~\citep{subramaniam2025multiagent}, sampling answers through Debate and optimizing each LLM using the original training algorithm. Each LLM will solve questions independently at inference time. We rename this baseline as DebateFT. 

\textbf{Training:} We use full-tuning to optimize the LLMs for one epoch. We use a batch size of $256$ and a learning rate of 1e-6 for Llama-3.1-8B-Instruct, 1e-6 for Qwen2.5-7B-Instruct, and 9e-7 for Qwen2.5-14B-Instruct. The temperature is 0.5 for all LLMs, and the KL coefficient is 0 for Llama-3.1-8B-Instruct, 5e-7 for Qwen2.5-7B-Instruct, and 0 for Qwen2.5-14B-Instruct. The maximum output token number of the sampled answer is 2K. As for other hyperparameters, we strictly use the original parameters of GRPO. 
We utilize Llama-3-8b-rm-mixture~\citep{hu2024openrlhf} as the reward model to rate sampled answers.
The initial reasoning ability $\gamma_{i}$, which can be measured on the validation set, is 0.59, 0.75, and 0.78 for Llama-3.1-8B-Instruct, Qwen2.5-7B-Instruct, and Qwen2.5-14B-Instruct. 
According to the IRT in Section~\ref{sec: dynamic interaction}, the proportion of cooperation/competition across the three groups is as follows: 36.30/63.70\%, 32.76/67.24\%, and 16.87/83.13\% for Group1, Group2, and Group3. 
The approximate training time is 8 hours for Group1, 12 hours for Group2, and 10 hours for Group3. 

\textbf{Evaluation:} We set the temperature as 0 and the maximum output token number of evaluation is set to 8K for AIME, while 2K for other benchmarks. All training experiments are conducted on eight H100 GPUs, and evaluation experiments are conducted on one H100 GPU.

\subsection{Statistics of Benchmarks}
\label{appendix: statistics of benchmarks}

\textbf{Training Dataset}

\begin{figure*}[h!]
\centering
\includegraphics[width=0.9\linewidth]{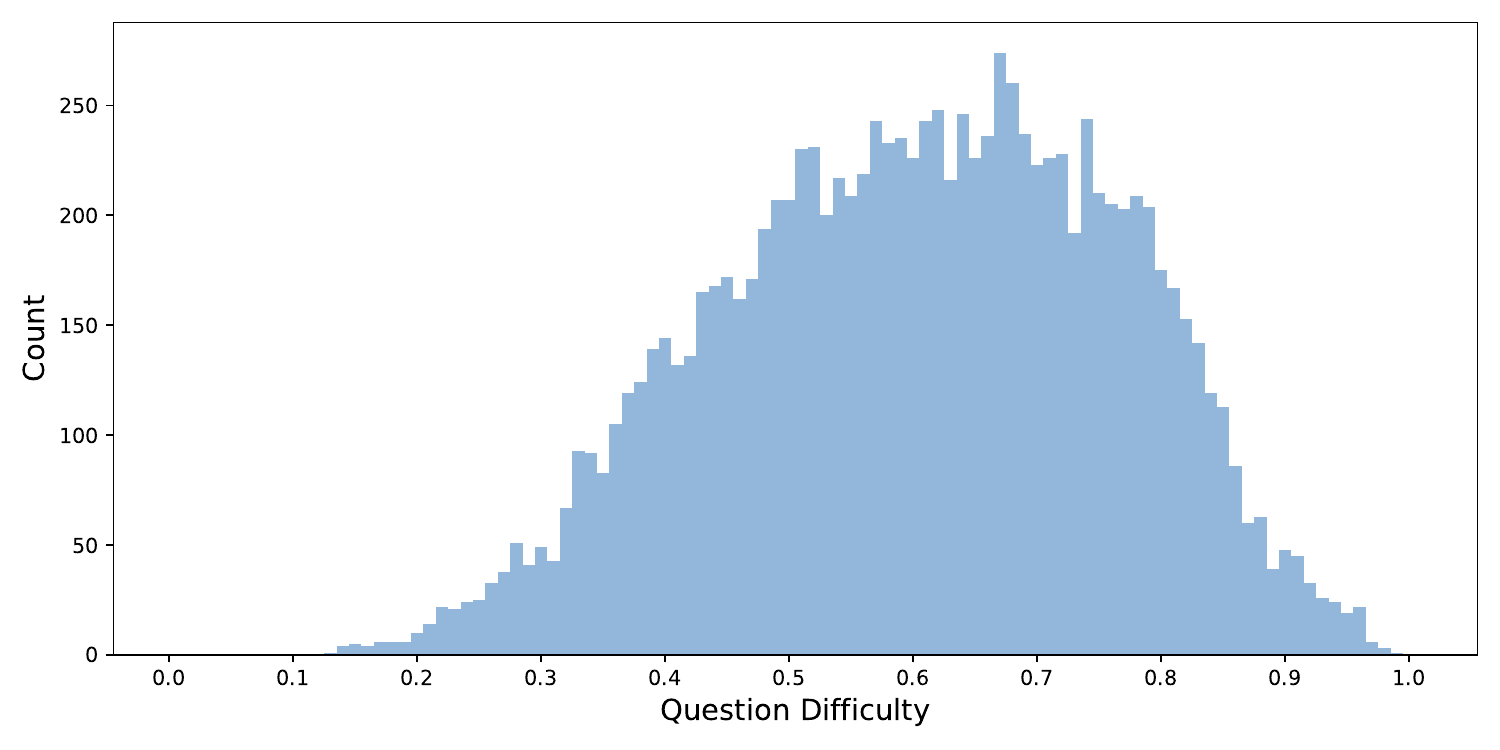}
\caption{Question Difficulty Distribution (the interval is 0.01) of MATH training set used in ILR.}
\label{fig: distribution}
\end{figure*}

\begin{table*}[h]
\centering
\begin{tabular}{@{}lccc@{}}
\toprule
Model  & ILR-Group1 & ILR-Group2 & ILR-Group3 \\ \midrule
Llama-3.1-8B-Instruct &5.04\%&7.15\%&-\\ 
Qwen2.5-7B-Instruct &2.19\%&-&1.64\%\\
Qwen2.5-14B-Instruct &-&3.59\%&2.34\%\\
\bottomrule
\end{tabular}%
\caption{The clip ratio of each group.}
\label{tab:clip_ratio}
\end{table*}

Following~\cite{zeng2025simplerl}, we only use MATH~\citep{hendrycks2021measuring} as our training data source. Excluding the common MATH-500~\citep{hendrycks2021measuring} as the evaluation set, there are 12000 samples in the remaining dataset. We randomly select 1000 samples as the validation set to assess the LLMs' reasoning ability for Dynamic Interaction, while the remaining 11000 samples serve as the training set to fine-tune LLMs.

Each question of the MATH training set will have a continuous difficulty measured by LLMs' self-ranking, and we depict the question difficulty distribution in Figure~\ref{fig: distribution}. As illustrated in Section~\ref{sec: experimental setting}, the initial reasoning ability $\gamma_{i}$, which can be measured on the validation set, is 0.59, 0.75, and 0.78 for Llama-3.1-8B-Instruct, Qwen2.5-7B-Instruct, and Qwen2.5-14B-Instruct. According to the IRT in Section~\ref{sec: dynamic interaction}, the proportion of cooperation/competition across the three groups is as follows: 36.30/63.70\%, 32.76/67.24\%, and 16.87/83.13\% for Group 1, Group 2, and Group 3, respectively.

\textbf{Evaluation Dataset}

\textbf{GSM8K}~\citep{cobbe2021training}: This dataset comprises 1319 single-step mathematical questions designed for elementary school students in English. As one of the most widely adopted benchmarks in the field, it plays a basic role in assessing the arithmetic reasoning capabilities of LLMs. 

\textbf{MATH-500}~\citep{hendrycks2021measuring}: The dataset encompasses 500 intermediate-level mathematical problems systematically curated across core domains, including algebra, geometry, precalculus, probability, and number theory. 

\textbf{Minerva Math}~\citep{lewkowycz2022solving}: This dataset contains 272 mathematical problems across core domains, including algebra, geometry, precalculus, probability, and number theory.

\textbf{Olympiad Bench}~\citep{he2024olympiadbench}: This benchmark constitutes a bilingual multimodal evaluation framework comprising 8476 Olympiad-level problems curated from prestigious mathematics and physics competitions. We utilize the subset processed by~\citep{yang2024qwen2} as our test set, which contains 675 English text-only questions.

\textbf{AIME24\&25}~\citep{aime}: This benchmark collection contains 60 questions and derives from the 2024 and 2025 editions of the American Invitational Mathematics Examination (AIME), comprising two distinct problem sets. Each set contains 30 rigorously vetted mathematical questions characterized by high cognitive demand. The evaluative focus lies in probing advanced mathematical competencies.

\textbf{MBPP}~\citep{austin2021program}: This benchmark comprises 974 crowd-sourced Python programming problems, meticulously curated to align with the competency level of entry-level programmers. These problems encompass foundational programming concepts, standard library implementations, and essential algorithmic paradigms, ensuring a comprehensive evaluation of introductory programming proficiency. Following~\citep{ni2023lever}, we evaluate ILR and baselines using the test subset, which contains 500 questions.

\camera{
\textbf{NQ}~\citep{kwiatkowski2019natural}: This benchmark is designed to evaluate open-domain question answering systems on natural, information-seeking questions that require reasoning over entire documents. We randomly sample 500 questions and evaluate ILR and baselines on this set.

\textbf{MMLU}~\citep{hendrycks2020measuring}: This comprehensive benchmark spans subjects across STEM, the humanities, and the social sciences, designed to measure a model's multitask accuracy and world knowledge. It includes questions ranging from elementary to professional levels, evaluating both knowledge acquisition and problem-solving abilities in zero-shot and few-shot settings. We randomly sample 500 questions and evaluate ILR and baselines on this set.
}

\revise{
\subsection{Clip ratio}
\label{clip ratio}
In Section~\ref{Perception Calibration}, we utilize a clip$(\cdot)$ function to stabilize the ILR training. In Table~\ref{tab:clip_ratio}, we report the proportion of rewards that are actually clipped. It shows that the percentage of instances where the normalized reward difference exceeded the boundary is small. Therefore, we believe that this clip will not overly suppress the informative signals.

}




\begin{table*}[t]

\centering
\begin{tabular}{@{}l|cc|cc@{}}
\toprule
Model & Avg & z-Avg & ratio & z-ratio \\ \midrule
Llama-3.1-8B-Instruct &\textbf{41.51}&40.74 ($\downarrow$ 0.77)&5.04&20.45 ($\uparrow$ 15.41)\\ 
Qwen2.5-7B-Instruct &\textbf{54.44}&53.94 ($\downarrow$ 0.50)&2.19&9.10 ($\uparrow$ 6.91)\\ 
\bottomrule
\end{tabular}%
\caption{\revise{Comparison with z-score normalization (z-) based on Group1. We report the average accuracy (\%) on five mathematical evaluation benchmarks and the clip ratio (\%).}}
\label{tab:z-score}
\end{table*}

\section{Discussion}
\label{appendix: discussion}
\subsection{Heterogeneous Role Assignments in ILR} 
Our ILR intentionally employs a homogeneous design where all agents are problem solvers. This choice is directly tied to our core research question: to investigate whether and how interaction within a multi-agent environment can enhance the independent problem-solving capabilities of each individual LLM. In this context, all agents share the same role and learn from each other through interaction to achieve mutual improvement. However, we believe the idea of incorporating heterogeneous roles is interesting, and an exciting future direction would be to design a hybrid system. For example, one could have multiple ``generator'' agents that are trained internally via ILR to enhance their respective independent generation capabilities. Concurrently, the system could include one or more ``verifier'' or ``refiner'' agents responsible for evaluating and integrating the outputs from these ILR-strengthened generators. Such a hybrid model could potentially lead to an even more complex, robust, and powerful multi-agent system.

\subsection{Robustness of Dynamic Interaction} 
For many common reasoning tasks, such as mathematics and standard question-answering, the intrinsic difficulty of a problem is one of the most direct and dominant factors influencing whether an individual chooses to persevere independently or seek collaborative assistance. However, in scenarios like creative writing, the ``problem difficulty'' does not necessarily correlate directly with the quality of the final output. Even a ``simple problem'' might benefit more from a cooperative strategy, as this could foster a richer blend of stylistic diversity by combining the unique perspectives of different LLMs. In such cases, the goal is not just to find a correct solution but to generate a high-quality, creative artifact.
Therefore, for such specific task settings, we need to incorporate additional conditioning variables to improve the robustness of Dynamic Interaction.

\section{More Experiment Results}
\label{appendix:more ablation study}
\subsection{Comparison with z-score normalization}
\label{appendix: z-score}
To investigate the effect of different normalization methods in reward shaping (Section~\ref{Perception Calibration}), we compare our min-max normalization with z-score normalization:
\begin{equation}
\begin{split}
    \delta_{i,l}=\frac{R_{i,k} - R_{l, avg}}{R_{l, std}}
\end{split}
\end{equation}

We report the average accuracy on five mathematical evaluation benchmarks and the clip ratio discussed in Appendix~\ref{clip ratio}.
Table~\ref{tab:z-score} shows that implementing z-score normalization within the ILR framework yields inferior performance compared to our method. We attribute this degradation to the aggressive clipping associated with z-score normalization, which tends to suppress the magnitude of incentive signals. This observation further corroborates the superiority of our min-max normalization.

\revise{
\subsection{Confidence Intervals of Table 1}
\label{appendix: confidence intervals}

In Table~\ref{tab:confidence}, we report the 95\% confidence intervals of Table~\ref{tab:mian_result}. Note that except for the result in Table~\ref{tab:mian_result} and Table~\ref{tab:confidence}, which are based on 5 random seeds, all the others are based on the result of seed 0.

}
\begin{table*}[h!]
\centering
\resizebox{0.8\textwidth}{!}{%
\begin{tabular}{@{}l|cccccc@{}}
\toprule
 & GSM8K & MATH-500 & \begin{tabular}[c]{@{}c@{}}Minerva\\ Math\end{tabular}& \begin{tabular}[c]{@{}c@{}}Olympiad\\ Bench\end{tabular} & \begin{tabular}[c]{@{}c@{}}AIME\\ 24\&25\end{tabular}  & Avg \\ \midrule
Llama-3.1-8B-Instruct &$\pm$1.78&$\pm$1.33&$\pm$1.79&$\pm$0.83&$\pm$1.13&$\pm$0.69 \\
SFT &$\pm$0.39&$\pm$1.61&$\pm$1.83&$\pm$1.12&$\pm$0.92&$\pm$0.68\\
DPO &$\pm$0.63&$\pm$1.48&$\pm$1.07&$\pm$0.94&$\pm$1.13&$\pm$0.18\\
PPO &$\pm$1.74&$\pm$1.62&$\pm$3.08&$\pm$0.93&$\pm$2.36&$\pm$0.93\\
GRPO &$\pm$0.72&$\pm$3.43&$\pm$1.82&$\pm$2.16&$\pm$2.36&$\pm$1.13\\
GSPO &$\pm$0.81&$\pm$0.32&$\pm$0.40&$\pm$0.25&$\pm$0.00&$\pm$0.34\\
Reinforce++ &$\pm$1.16&$\pm$2.66&$\pm$1.65&$\pm$1.71&$\pm$2.27&$\pm$0.72\\
DebateFT-Group1 &$\pm$1.78&$\pm$3.01&$\pm$1.73&$\pm$2.40&$\pm$1.85&$\pm$1.85\\
DebateFT-Group2 &$\pm$0.73&$\pm$2.72&$\pm$4.32&$\pm$3.35&$\pm$1.85&$\pm$2.03\\ 
\rowcolor{lightgreen}
ILR-Group1 &$\pm$0.70&$\pm$0.97&$\pm$0.24&$\pm$0.18&$\pm$1.14&$\pm$0.55\\
\rowcolor{lightgreen}
ILR-Group2 &$\pm$0.21&$\pm$1.14&$\pm$1.18&$\pm$0.71&$\pm$1.46&$\pm$0.77\\ \midrule

Qwen2.5-7B-Instruct &$\pm$0.57&$\pm$1.35&$\pm$1.84&$\pm$0.62&$\pm$2.78&$\pm$1.20\\
SFT &$\pm$0.86&$\pm$2.10&$\pm$1.63&$\pm$1.44&$\pm$3.70&$\pm$1.43\\
DPO &$\pm$0.62&$\pm$2.02&$\pm$1.50&$\pm$0.33&$\pm$3.40&$\pm$1.10\\
PPO &$\pm$0.83&$\pm$1.39&$\pm$2.03&$\pm$0.82&$\pm$2.07&$\pm$1.23\\
GRPO&$\pm$0.39&$\pm$1.12&$\pm$1.12&$\pm$1.05&$\pm$2.70&$\pm$0.97\\
GSPO&$\pm$0.31&$\pm$1.65&$\pm$0.62&$\pm$0.93&$\pm$0.93&$\pm$0.82\\
Reinforce++ &$\pm$0.24&$\pm$0.88&$\pm$1.14&$\pm$0.53&$\pm$1.85&$\pm$0.80\\
DebateFT-Group1 &$\pm$0.54&$\pm$1.72&$\pm$2.88&$\pm$1.37&$\pm$0.93&$\pm$1.10\\
DebateFT-Group3 &$\pm$0.41&$\pm$0.59&$\pm$3.03&$\pm$1.08&$\pm$1.85&$\pm$0.86\\ 
\rowcolor{lightgreen}
ILR-Group1 &$\pm$0.31&$\pm$0.46&$\pm$1.18&$\pm$0.28&$\pm$2.26&$\pm$0.58\\
\rowcolor{lightgreen}
ILR-Group3 &$\pm$0.20&$\pm$0.25&$\pm$0.68&$\pm$0.28&$\pm$2.26&$\pm$0.48\\ \midrule

Qwen2.5-14B-Instruct &$\pm$0.18&$\pm$1.45&$\pm$2.32&$\pm$1.18&$\pm$3.76&$\pm$1.13\\
SFT &$\pm$0.28&$\pm$1.02&$\pm$2.11&$\pm$1.17&$\pm$3.07&$\pm$1.23\\
DPO &$\pm$0.27&$\pm$0.79&$\pm$2.54&$\pm$2.39&$\pm$1.14&$\pm$0.95\\
PPO &$\pm$0.49&$\pm$1.33&$\pm$1.84&$\pm$1.91&$\pm$2.53&$\pm$1.43\\
GRPO&$\pm$0.72&$\pm$1.50&$\pm$2.28&$\pm$1.13&$\pm$3.14&$\pm$1.47\\
GSPO&$\pm$0.17&$\pm$0.42&$\pm$0.38&$\pm$0.19&$\pm$0.93&$\pm$0.36\\
Reinforce++ &$\pm$0.25&$\pm$1.22&$\pm$0.95&$\pm$1.79&$\pm$4.86&$\pm$1.42\\
DebateFT-Group2 &$\pm$0.18&$\pm$0.97&$\pm$2.05&$\pm$1.11&$\pm$1.73&$\pm$0.30\\
DebateFT-Group3 &$\pm$0.56&$\pm$0.75&$\pm$2.63&$\pm$1.25&$\pm$3.70&$\pm$1.41\\ 
\rowcolor{lightgreen}
ILR-Group2 &$\pm$0.08&$\pm$0.21&$\pm$0.41&$\pm$0.43&$\pm$2.53&$\pm$0.72\\
\rowcolor{lightgreen}
ILR-Group3 &$\pm$0.14&$\pm$0.33&$\pm$0.99&$\pm$0.51&$\pm$2.70&$\pm$0.86\\ \bottomrule
\end{tabular}%
}
\caption{\revise{95\% Confidence Intervals of Table 1.}}
\label{tab:confidence}
\end{table*}

\begin{table*}[t]

\centering
\begin{tabular}{@{}l|cccccc@{}}
\toprule
 & GSM8K & MATH-500 & \begin{tabular}[c]{@{}c@{}}Minerva\\ Math\end{tabular}& \begin{tabular}[c]{@{}c@{}}Olympiad\\ Bench\end{tabular} & \begin{tabular}[c]{@{}c@{}}AIME\\ 24\&25\end{tabular}  & Avg \\ \midrule
Llama-3.1-8B-Instruct &82.87&49.80&22.79&13.63&1.67&34.15\\
Re-ReST &84.23&49.80&29.04&15.56&5.00&36.73\\
CPPO &84.84&54.60&27.57&18.85&6.67&38.51\\ 
2-GRPO &86.43&53.20&28.31&18.22&6.67&38.57\\ 
ILR-Group1 &\textbf{89.39}&\textbf{55.80}&30.15&22.22&\textbf{10.00}&\textbf{41.51}\\
ILR-Group2 &87.26&55.20&\textbf{33.82}&\textbf{22.52}&6.67&41.10\\ \bottomrule
\end{tabular}%
\caption{\revise{Comparison with Re-ReST, CPPO, and 2-GRPO, on Llama-3.1-8B-Instruct.}}
\label{tab:more_self_learning}
\end{table*}

\begin{table*}[t]

\centering
\begin{tabular}{@{}l|cc|cc@{}}
\toprule
Model & Avg & periodic-Avg & proportion & periodic-proportion \\ \midrule
Llama-3.1-8B-Instruct &41.51&\textbf{42.35} ($\uparrow$ 0.84)&\multirow{2}{*}{36.30/63.70\%}&\multirow{2}{*}{32.44/67.56\%}\\ 
Qwen2.5-7B-Instruct &54.44&\textbf{54.82} ($\uparrow$ 0.38)&&\\ 
\bottomrule
\end{tabular}%
\caption{\revise{Comparison with periodic updates of LLM's reasoning ability $\gamma_i$ (periodic-) based on Group1. We report the average accuracy on five mathematical evaluation benchmarks and the proportion of cooperation/competition.}}
\label{tab:periodic updates}
\end{table*}

\begin{table*}[h!]
\centering
\begin{tabular}{@{}l|cc|cc@{}}
\toprule
Model & Avg & mixed-Avg & proportion & mixed-proportion \\ \midrule
Llama-3.1-8B-Instruct &41.51&\textbf{41.91} ($\uparrow$ 0.40)&\multirow{2}{*}{36.30/63.70\%}&\multirow{2}{*}{40.99/59.01\%}\\ 
Qwen2.5-7B-Instruct &\textbf{54.44}&53.97 ($\downarrow$ 0.47)&&\\ 
\bottomrule
\end{tabular}%
\caption{\revise{Comparison with mixed types interaction (mixed-) based on Group1. We report the average accuracy on five mathematical evaluation benchmarks and the proportion of cooperation/competition.}}
\label{tab:mixed types}
\end{table*}

\begin{table*}[t]

\centering
\begin{tabular}{@{}l|cccc@{}}
\toprule
Model & GRPO & ILR-Group1 & ILR- Group2 & ILR-Group3 \\ \midrule
Llama-3.1-8B-Instruct &38.39[3.75]&41.51[8.00]&41.10[12.00]&-\\ 
Qwen2.5-7B-Instruct &53.58[3.25]&54.44[8.00]&-&54.59[10.00]\\ 
Qwen2.5-14B-Instruct &57.55[5.50]&-&58.95[12.00]&59.30[10.00]\\ 
\bottomrule
\end{tabular}%
\caption{\revise{Quantitative analysis of computational overhead versus performance gains. We report the average accuracy on five mathematical evaluation benchmarks (\%, without []) and the training time (hours, within []).}}
\label{tab:cost_table}
\end{table*}

\begin{table*}[h!]

\centering
\begin{tabular}{@{}l|cccccc@{}}
\toprule
& GSM8K & MATH-500 & \begin{tabular}[c]{@{}c@{}}Minerva\\ Math\end{tabular}& \begin{tabular}[c]{@{}c@{}}Olympiad\\ Bench\end{tabular} & \begin{tabular}[c]{@{}c@{}}AIME\\ 24\&25\end{tabular}  & Avg \\ \midrule
Llama-3.1-8B-Instruct  &82.87&49.80&22.79&13.63&1.67&34.15 \\
Prompt1 &87.79&54.20&29.41&20.30&6.67&39.67\\ 
Prompt2 &87.72&55.60&30.51&20.15&6.67&40.13\\ \midrule
ILR-Group1 &\textbf{89.39}&\textbf{55.80}&30.15&22.22&\textbf{10.00}&\textbf{41.51} \\ 
ILR-Group2 &87.26&55.20&\textbf{33.82}&\textbf{22.52}&6.67&41.10\\
\bottomrule
\end{tabular}%
\caption{\revise{Comparison of the same LLM with different prompts based on Llama-3.1-8B-Instruct.}}
\label{tab:same_llm_different_prompts}
\end{table*}

\begin{table*}[h!]
\centering
\resizebox{0.9\textwidth}{!}{%
\begin{tabular}{@{}l|c|cccccc@{}}
\toprule
 & \begin{tabular}[c]{@{}c@{}}Optimization\\ algorithms\end{tabular} & GSM8K & \begin{tabular}[c]{@{}c@{}}MATH\\ -500\end{tabular} & \begin{tabular}[c]{@{}c@{}}Minerva\\ Math\end{tabular}& \begin{tabular}[c]{@{}c@{}}Olympiad\\ Bench\end{tabular} & \begin{tabular}[c]{@{}c@{}}AIME\\ 24\&25\end{tabular}  & Avg \\ \midrule
Llama-3.1-8B-Instruct &-&82.87&49.80&22.79&13.63&1.67&34.15\\
\rowcolor{lightgreen}
ILR-Group1 &GRPO&89.39&55.80&30.15&22.22&10.00&41.51\\
\rowcolor{lightgreen}
ILR-Group2 &GRPO&87.26&55.20&33.82&22.52&6.67&41.10\\ 
\rowcolor{lightpink}
ILR-Group1 &GSPO&89.72&56.00&31.25&21.78&10.00&41.75$\uparrow$\\
\rowcolor{lightpink}
ILR-Group2 &GSPO&89.58&54.80&30.15&23.15&10.00&41.54$\uparrow$\\ 
\midrule

Qwen2.5-7B-Instruct &-&92.34&75.60&41.54&37.63&11.67&51.76\\
\rowcolor{lightgreen}
ILR-Group1 &GRPO&93.40&77.60&43.01&39.85&18.33&54.44\\
\rowcolor{lightgreen}
ILR-Group3 &GRPO&92.65&78.00&45.01&38.96&18.33&54.59\\ 
\rowcolor{lightpink}
ILR-Group1 &GSPO&93.48&77.80&45.22&40.15&16.67&54.66$\uparrow$\\
\rowcolor{lightpink}
ILR-Group3 &GSPO&92.95&78.60&44.85&40.30&20.00&55.34$\uparrow$\\ 
\midrule

Qwen2.5-14B-Instruct &-&94.84&81.20&47.43&41.04&13.33&55.57\\
\rowcolor{lightgreen}
ILR-Group2 &GRPO&95.53&81.80&50.37&43.70&23.33&58.95\\
\rowcolor{lightgreen}
ILR-Group3 &GRPO&95.30&82.60&51.10&44.15&23.33&59.30\\ 
\rowcolor{lightpink}
ILR-Group2 &GSPO&95.98&82.80&50.37&43.70&23.33&59.24$\uparrow$\\
\rowcolor{lightpink}
ILR-Group3 &GSPO&95.75&82.20&52.37&43.96&23.33&59.52$\uparrow$\\ 
\bottomrule
\end{tabular}%
}
\caption{Quantitative comparison of using different reinforcement learning algorithms (GRPO vs. GSPO) as the ILR optimization algorithms. Green lines are based on GRPO, and pink lines are based on GSPO.}
\label{tab:gspo}
\end{table*}

\revise{
\subsection{Comparison with More Self-learning Methods}
\label{appendix: more self-learning methods}
In Table~\ref{tab:mian_result}, we compare ILR with several basic self-learning methods like SFT and GRPO. Here, we discuss more self-learning baselines and compare ILR with them based on Llama-3.1-8B-Instruct.


\textbf{Self-reflection Learning.} Self-reflection and self-improvement constitute a pivotal research trajectory for enhancing agent capabilities through an intra-agent feedback loop~\citep{renze2024self, shinn2023reflexion, dou2024re, zhang2024self, he2025self}. The central premise is that an agent can refine its performance by critically analyzing its own generated trajectories, answers, or reasoning processes in a closed-loop manner. To rigorously benchmark our multi-agent approach against this single-agent paradigm, we employ Re-ReST as a representative baseline. This comparison aims to determine whether the diverse feedback from peers offers superior gradients for improvement compared to the internalized self-verification.


\textbf{Advanced GRPO Variants.} Recent advancements in reinforcement learning have introduced sophisticated variants of GRPO~\citep{lin2025cppo, sun2025cog}, which significantly outperform the vanilla GRPO implementation. By comparing against CPPO~\citep{lin2025cppo}, we seek to unveil whether the multi-agent enhancement provides a distinct advantage over purely algorithmic improvements in single-agent training stability.


\textbf{GRPO with More Sampling (Compute-Matched Baseline).} To rule out the possibility that performance gains are simply artifacts of increased sampling, we introduce a computation-aligned baseline: single-agent GRPO with doubled rollout numbers (denoted as 2-GRPO). Since our multi-agent setting involves interactions that effectively increase the inference load, this variant establishes a fair comparison by equating the sampling volume. By contrasting ILR with this high-sampling baseline, we aim to demonstrate that the benefits stem from the quality of interactive feedback rather than from more sampling.

\highlight{
The results presented in Table~\ref{tab:more_self_learning} demonstrate that ILR consistently outperforms the extended suite of self-learning baselines, substantiating the distinct advantage of our multi-agent framework. Notably, ILR surpasses the self-reflection method Re-ReST, suggesting that diverse feedback from peer agents effectively mitigates the inherent blind spots of single-agent introspection. Moreover, ILR demonstrates its superiority over CPPO, underscoring that the benefits of multi-agent enhancement transcend purely algorithmic improvements in single-agent stability. Importantly, ILR also outperforms the computation-aligned 2-GRPO baseline, which confirms that the effectiveness of ILR stems from the high quality of interactive rather than the sheer quantity of samples, demonstrating greater data efficiency than force-scaling of rollouts.
}
}

\revise{
\subsection{Comparison with periodic updates of LLM's reasoning ability}
\label{appendix: periodic}
As a model's reasoning ability can indeed evolve during training. To investigate this, we conduct an additional experiment to study the impact of dynamically updating $\gamma_i$. Specifically, we re-evaluated $\gamma_i$ periodically every 5 training steps (we have a total of 42 steps). We report the average accuracy on five mathematical evaluation benchmarks and the proportion of cooperation/competition. Table~\ref{tab:periodic updates} indicates that the performance of periodic-updates surpasses that of one-time measurement. As the capability gap diminishes, the cooperation rate declines, a dynamic that appears more organic, thereby further enhancing performance.

}

\subsection{Comparison with mixed types interaction}
\label{appendix: mixed types interaction}
In real human collaboration, cooperation and competition often coexist within the same reasoning process. To investigate this scenario, we modified our sampling procedure as follows: for a given problem, even when our IRT-based mechanism determines to use a cooperation strategy, we enforce a mixture of strategies across the multiple samples. For example, if we generated five distinct rollouts for a single question, we would randomly assign the designated cooperation strategy to 3-5 of these rollouts, while assigning the competition strategy to the remaining rollouts. 

As shown in Table~\ref{tab:mixed types}, this shift yielded asymmetric outcomes. Llama-3.1-8B-Instruct achieved a performance gain of 0.40\%, likely because the mixed strategy exposed it to more cooperative reasoning paths on challenging problems. In contrast, Qwen2.5-7B-Instruct experienced a regression of 0.47\%. This decline suggests that the ``mixed'' samples acted as noise for the stronger model, where forcing it to collaborate (when it should compete) likely constrained its reasoning potential or introduced errors from the weaker partner.

\revise{
\subsection{Cost Table}
\label{appendix: cost}
We compare the training time and performance of GRPO and ILR in Table~\ref{tab:cost_table}.
Note that the training durations reported for ILR-Group$i$ represent the aggregated time required for both LLMs within the group. For instance, in the ILR-Group1 setting, the combined training time for Llama-3.1-8B-Instruct and Qwen2.5-7B-Instruct amounts to 8 hours. In comparison, training these models individually using standard GRPO requires a total of 7 hours (3.75 hours for Llama-3.1-8B-Instruct and 3.25 hours for Qwen2.5-7B-Instruct). Consequently, incorporating ILR introduces only a marginal computational overhead of one hour. This slight increase in cost is justified by the performance gains, yielding improvements of approximately 3\% for Llama-3.1-8B-Instruct and 1\% for Qwen2.5-7B-Instruct.

}

\revise{
\subsection{Same LLM with different input prompts}
\label{appendix: same llm different prompts}
In our primary experiments, we conceptualized distinct LLMs receiving the same input prompt as individual agents to investigate interactive learning among heterogeneous models. In this section, we extend this inquiry by examining the efficacy of interactive learning within a homogeneous setting, where a single LLM utilizes different input prompts to simulate distinct agents.

\begin{itemize}
    \item \textbf{Prompt1}: ${<Question>}$ Please reason step by step.
    \item \textbf{Prompt2}: ${<Question>}$ Please double-check the question, first extract the keywords and think about the knowledge about these keywords, then solve the question.
\end{itemize}

Two notable findings emerge from this analysis (see Table~\ref{tab:same_llm_different_prompts}). \textbf{First}, while ILR with an identical LLM still yields performance improvements over single-agent baselines, it does not surpass the performance of ILR with different LLMs. This suggests that while diversity can be simulated through varied prompting, distinct models provide a richer and more effective source of alternative perspectives. The interaction between truly heterogeneous agents appears to generate a more powerful learning signal. \textbf{Second}, the post-training performance of the Llama model utilizing Prompt2 was superior to its counterpart. This observation underscores that the granularity of instruction prompts plays a significant role in influencing the outcomes of interactive learning.
}

\subsection{Extendability of Optimization Algorithm}
\label{appendix: extendability}
In fact, the optimization algorithm used in our ILR is not limited to GRPO and can be replaced by another method. Any reinforcement learning (RL) algorithm capable of generating a group of responses can serve as our optimization algorithm. Therefore, in this section, we investigate the impact of different optimization algorithms (GRPO vs. GSPO) on ILR. As shown in the baseline results of Table~\ref{tab:mian_result}, GSPO outperforms GRPO across all LLMs. Meanwhile, from Table~\ref{tab:gspo}, it can be observed that ILR with GSPO achieves better results than ILR using GRPO. This demonstrates the high extensibility and universality of ILR with respect to RL algorithms. With the advancement of more sophisticated RL algorithms in the future, ILR is expected to achieve even stronger performance.

\revise{
\section{Pseudocode of ILR}
\label{appendix: pseudocode}

\begin{algorithm}[H] 
\caption{Interactive Learning for LLM Reasoning (ILR Training Algorithm)} 
\label{alg:ilr}
\footnotesize 
\begin{algorithmic}[1]
\State \textbf{Input:} Training set $\mathcal{D} = \{q_1, \dots, q_N\}$, Validation set $\mathcal{D}_{val}$, Set of LLMs $\mathcal{M} = \{M_1, \dots, M_m\}$, Reward model $\mathcal{M}_{reward}$, Self-ranking splits $S$, Batch size $N'$, Sampling rounds $n$
\State \textbf{Output:} Optimized Models $\mathcal{M}^*$

\vspace{10pt}
\Statex \textbf{\textit{Phase 1: Ability and Difficulty Estimation}}
\State Estimate model ability $\gamma_i$ for each $M_i \in \mathcal{M}$ based on accuracy on $\mathcal{D}_{val}$
\For{each question $q \in \mathcal{D}$}
    \State Divide $\mathcal{D}$ into random batches of size $N'$ across $S$ splits
    \State Collect ranks $r_{q,j}$ from $M_i$ via self-ranking prompt
    \State $D_q \leftarrow \frac{1}{m} \sum_{i=1}^{m} M_i \big(\frac{1}{S} \sum_{j=1}^{S} \frac{r_{q,j}}{N'}\big)$  \Comment{Eq. 1}
\EndFor

\vspace{10pt}
\Statex \textbf{\textit{Phase 2: Training Loop}}
\For{each batch $\mathcal{B} \subset \mathcal{D}$}
    \State $\mathcal{T} \leftarrow \emptyset$, $\mathcal{R} \leftarrow \emptyset$, $\bar{\mathcal{R}} \leftarrow \emptyset$
    \For{each question $q \in \mathcal{B}$}
        \vspace{5pt}
        \State \textit{// 1. Dynamic Interaction}
        \State Calculate success prob.
        \State $P_{q,i} \leftarrow \frac{1}{1 + e^{-1.7(\gamma_i - D_q)}}$ for all $M_i$ \Comment{Eq. 2}
        \State $P_q \leftarrow \frac{1}{m} \sum_{i=1}^{m} P_{q,i}$ 
        \If{$P_q < 0.5$} $\text{Mode} \leftarrow \text{Cooperation}$ \Else \ $\text{Mode} \leftarrow \text{Competition}$ \EndIf \Comment{Eq. 3}

        \vspace{5pt}
        \State \textit{// 2. Idea3 Interaction}
        \For{$k = 1$ to $n$, $i = 1$ to $m$}
            \State $a_{init} \leftarrow M_i.\text{Gen}(q, \text{Prompt}_{\text{S}})$
            \State $a_{ana} \leftarrow M_i.\text{Gen}(q, a_{init}^{\text{peer}}, \text{Mode}, \text{Prompt}_{\text{A}})$
            \State $a_{fus} \leftarrow M_i.\text{Gen}(q, a_{ana}, \text{Mode}, \text{Prompt}_{\text{F}})$
            
            \If{$\text{IsCorrect}(a_{init}) \land \neg \text{IsCorrect}(a_{fus})$}
                    \State $a_{final} \leftarrow a_{init}$
                \Else
                    \State $a_{final} \leftarrow a_{fus}$
                \EndIf
            
            \State $\tau_{i,k} \leftarrow (q, a_{final})$; $\mathcal{T}.\text{add}(\tau_{i,k})$
            \State $R_{i,k} \leftarrow \mathcal{M}_{r}(\tau_{i,k})$; $\mathcal{R}.\text{add}(R_{i,k})$
        \EndFor

        \vspace{5pt}
        \State \textit{// 3. Perception Calibration}
        \For{$i = 1$ to $m$}
            \For{$k = 1$ to $n$}
                \State $V_{calib} \leftarrow 0$
                \For{$l \in \mathcal{M} \setminus \{M_i\}$}
                    \State Get peer stats $R_{l,max}, R_{l,min}, R_{l,avg}$
                    \State $\Delta \leftarrow (R_{i,k} - R_{l,avg}) / (R_{l,max} - R_{l,min})$
                    \State $V_{calib} \mathrel{+}= \text{clip}(\Delta, -\frac{1}{m-1}, +\frac{1}{m-1})$
                \EndFor
                \State $\bar{R}_{i,k} \leftarrow R_{i,k} + V_{calib}$ \Comment{Eq. 4}
                \State $\bar{\mathcal{R}}.\text{add}(\bar{R}_{i,k})$
            \EndFor
        \EndFor
    \EndFor

    \vspace{5pt}
    \State \textit{// 4. Optimization}
    \For{$i = 1$ to $m$}
        \State Update $M_i$ via GRPO using trajectories $\mathcal{T}$ and calibrated rewards $\bar{\mathcal{R}}$
    \EndFor
\EndFor
\end{algorithmic}
\end{algorithm}
}

\section{Case Study}
\label{appendix: case study}
We show some cases of \textbf{Dynamic Interaction (Idea3)} based on Group1 (LLM1: Llama-3.1-8B-Instruct and LLM2: Qwen2.5-7B-Instruct).

\begin{tcolorbox}[
  enhanced,
  breakable,
  fonttitle = \small\bfseries, 
  title=Cooperation mode (both LLMs initially fail but succeed after interaction),
  colframe=Salmon!90!Black,   
  colback=Salmon!20,     
  boxrule=1pt,
  boxsep=2pt,
  left=5pt,
  right=5pt,
  fontupper=\footnotesize,
  halign title = flush center
]
\textbf{Problem:}

The function $f(x)$ satisfies \(f(x) + f \left( \frac{x + 1}{1 - 3x} \right) = x\) for all $x \neq \frac{1}{3}.$  Find $f(2).$

\textbf{LLM1 (Idea Sharing):}

To find $f(2),$ we can use the given functional equation and plug in $x = 2.$  This gives us\[f(2) + f \left( \frac{2 + 1}{1 - 3 \cdot 2} \right) = 2 \Rightarrow f(2) + f \left( -\frac{3}{5} \right) = 2\]Now, we can plug in $x = -\frac{3}{5}$ into the original functional equation to get\[f \left( -\frac{3}{5} \right) + f \left( \frac{-\frac{3}{5} + 1}{1 - 3 \cdot \left( -\frac{3}{5} \right)} \right) = -\frac{3}{5}\]  
\[\Rightarrow f \left( -\frac{3}{5} \right) + f \left( \frac{1}{7} \right) = -\frac{3}{5}\]Now, we can plug in $x = \frac{1}{7}$ into the original functional equation to get\[f \left( \frac{1}{7} \right) + f \left( \frac{\frac{1}{7} + 1}{1 - 3 \cdot \frac{1}{7}} \right) = \frac{1}{7} \Rightarrow f \left( \frac{1}{7} \right) + f(2) = \frac{1}{7}\]Now we have two equations and two unknowns.  We can solve for $f(2)$ by subtracting the second equation from the first equation:\[f(2) + f \left( -\frac{3}{5} \right) - \left( f \left( \frac{1}{7} \right) + f(2) \right) = 2 - \frac{1}{7}\] \[ \Rightarrow f \left( -\frac{3}{5} \right) - f \left( \frac{1}{7} \right) = \frac{13}{7}\]Now we can plug in $f \left( -\frac{3}{5} \right) = 2 - f \left( \frac{1}{7} \right)$ into the equation $f \left( -\frac{3}{5} \right) + f \left( \frac{1}{7} \right) = -\frac{3}{5}$ to get\[2 - f \left( \frac{1}{7} \right) - f \left( \frac{1}{7} \right) = -\frac{3}{5} \Rightarrow f \left( \frac{1}{7} \right) = \frac{13}{10}\]Now we can plug in $f \left( \frac{1}{7} \right) = \frac{13}{10}$ into the equation $f \left( -\frac{3}{5} \right) - f \left( \frac{1}{7} \right) = \frac{13}{7}$ to get\[f \left( -\frac{3}{5} \right) - \frac{13}{10} = \frac{13}{7} \Rightarrow f \left( -\frac{3}{5} \right) = \frac{221}{70}\]Now we can plug in $f \left( -\frac{3}{5} \right) = \frac{221}{70}$ into the equation $f(2) + f \left( -\frac{3}{5} \right) = 2$ to get\[f(2) + \frac{221}{70} = 2 \Rightarrow f(2) = -\frac{81}{70}\]The final answer is: $\boxed{-\frac{81}{70}}$. \textcolor{red}{$\times$}

\textbf{LLM2 (Idea Sharing):}

To solve for \( f(2) \) given the functional equation\[ f(x) + f \left( \frac{x + 1}{1 - 3x} \right) = x \] We need to find a way to express \( f(x) \) in a form that allows us to determine \( f(2) \). First, let's denote the transformation \( g(x) = \frac{x + 1}{1 - 3x} \). We need to find the inverse of \( g(x) \) to understand the behavior of the function under repeated applications. To find the inverse, we solve for \( x \) in terms of \( y \) where \( y = g(x) \):\[ y = \frac{x + 1}{1 - 3x} \Rightarrow x = \frac{y - 1}{1 + 3y} \] Thus, the inverse function is\[ g^{-1}(x) = \frac{x - 1}{1 + 3x} \] Now, we apply the functional equation twice. First, we substitute \( x \) with \( g(x) \):\[ f(g(x)) + f \left( \frac{g(x) + 1}{1 - 3g(x)} \right) = g(x) \] Since \( g \left( \frac{x + 1}{1 - 3x} \right) = x \), we have:\[ f(g(x)) + f(x) = g(x) \] Next, we substitute \( x \) with \( g^{-1}(x) \):\[ f(g^{-1}(x)) + f \left( \frac{g^{-1}(x) + 1}{1 - 3g^{-1}(x)} \right) = g^{-1}(x) \] Since \( g \left( \frac{x - 1}{1 + 3x} \right) = x \), we have:\[ f(g^{-1}(x)) + f(x) = g^{-1}(x) \] We now have two equations: 1. \( f(g(x)) + f(x) = g(x) \) 2. \( f(g^{-1}(x)) + f(x) = g^{-1}(x) \). To find \( f(2) \), we need to determine the value of \( x \) such that \( g(x) = 2 \):\[ 2 = \frac{x + 1}{1 - 3x} \Rightarrow x = \frac{1}{7} \]Thus, \( g \left( \frac{1}{7} \right) = 2 \). Using the first equation with \( x = \frac{1}{7} \):\[ f(2) + f \left( \frac{1}{7} \right) = 2 \] Next, we need to find \( f \left( \frac{1}{7} \right) \). We use the second equation with \( x = 2 \):\[ f \left( \frac{2 - 1}{1 + 3 \cdot 2} \right) + f(2) = \frac{2 - 1}{1 + 3 \cdot 2}\] \[ \Rightarrow f \left( \frac{1}{7} \right) + f(2) = \frac{1}{7} \] We now have the system of linear equations: \( f(2) + f \left( \frac{1}{7} \right) = 2 \) and \( f \left( \frac{1}{7} \right) + f(2) = \frac{1}{7}. \) This is a contradiction, indicating an error in the setup. Revisiting the problem, we find:\( f(2) = \frac{2}{3} \) Thus, the correct value is:$\boxed{\frac{2}{3}}$. \textcolor{red}{$\times$}

\textbf{LLM1 (Idea Analysis):}

The partner's contribution provides a clear and step-by-step approach to solving the problem. The key steps that can help improve the original answer are: 1. Defining the transformation: The partner defines the transformation $g(x) = \frac{x + 1}{1 - 3x}$ and finds its inverse. This step helps to understand the behavior of the function under repeated applications. 2. Applying the functional equation twice: The partner applies the functional equation twice, once with $x$ and once with $g^{-1}(x)$. This step leads to two equations that can be used to solve for $f(2)$. This step is similar to the step-by-step plug-in solution. 3. Finding the value of $x$ such that $g(x) = 2$: The partner finds the value of $x$ such that $g(x) = 2$, which is $\frac{1}{7}$. This step is crucial for using the first equation to solve for $f(2)$. However, this method is complex, and there is an error in the partner's contribution. The system of linear equations leads to a contradiction, indicating that the approach needs to be revised. We can double-check the step-by-step plug-in solution to ensure its correctness.

\textbf{LLM2 (Idea Analysis):}

Let's analyze the key steps in the partner's contribution and identify areas for improvement. Key Steps in the Partner's Contribution: 1. Substitute \( x = 2 \) into the functional equation. 2. Substitute \( x = -\frac{3}{5} \) into the functional equation. 3. Substitute \( x = \frac{1}{7} \) into the functional equation. 4. Solve the system of equations. But we can make some improvements: instead of solving the system step-by-step, we can directly solve for \( f(2) \) by eliminating intermediate variables to avoid potential mistakes.

\textbf{LLM1 (Idea Fusion):}

To find $f(2),$ we can use the given functional equation and plug in $x = 2.$  This gives us\[f(2) + f \left( \frac{2 + 1}{1 - 3 \cdot 2} \right) = 2 \Rightarrow f(2) + f \left( -\frac{3}{5} \right) = 2\] Now, we can plug in $x = -\frac{3}{5}$ into the original functional equation to get\[f \left( -\frac{3}{5} \right) + f \left( \frac{-\frac{3}{5} + 1}{1 - 3 \cdot \left( -\frac{3}{5} \right)} \right) = -\frac{3}{5} \]\[\Rightarrow f \left( -\frac{3}{5} \right) + f \left( \frac{1}{7} \right) = -\frac{3}{5}\] Now, we can plug in $x = \frac{1}{7}$ into the original functional equation to get\[f \left( \frac{1}{7} \right) + f \left( \frac{\frac{1}{7} + 1}{1 - 3 \cdot \frac{1}{7}} \right) = \frac{1}{7} \]\[\Rightarrow f \left( \frac{1}{7} \right) + f(2) = \frac{1}{7}\] Now we have two equations and two unknowns. We can solve for $f(2)$ by adding the two equations:\[f(2) + f \left( -\frac{3}{5} \right) + f \left( \frac{1}{7} \right) + f(2) = 2 + \frac{1}{7} \]\[\Rightarrow 2f(2) + f \left( -\frac{3}{5} \right) + f \left( \frac{1}{7} \right) = \frac{15}{7}\] Now we can plug in $f \left( -\frac{3}{5} \right) + f \left( \frac{1}{7} \right) = -\frac{3}{5}$ into the equation to get\[2f(2) - \frac{3}{5} = \frac{15}{7} \Rightarrow f(2) = \frac{48}{35}\]The final answer is: $\boxed{\frac{48}{35}}$. \textcolor{blue}{$\checkmark$}

\textbf{LLM2 (Idea Fusion):}

To solve for \( f(2) \) given the functional equation\[ f(x) + f \left( \frac{x + 1}{1 - 3x} \right) = x \] We will follow a systematic approach. 1. Substitute \( x = 2 \) into the functional equation:   \[   f(2) + f \left( \frac{2 + 1}{1 - 3 \cdot 2} \right) = 2 \Rightarrow f(2) + f \left( -\frac{3}{5} \right) = 2   \] 2. Substitute \( x = -\frac{3}{5} \) into the functional equation:   \[   f \left( -\frac{3}{5} \right) + f \left( \frac{-\frac{3}{5} + 1}{1 - 3 \cdot \left( -\frac{3}{5} \right)} \right) = -\frac{3}{5} \]\[\Rightarrow  f \left( -\frac{3}{5} \right) + f \left( \frac{1}{7} \right) = -\frac{3}{5}   \] 3. Substitute \( x = \frac{1}{7} \) into the functional equation:   \[   f \left( \frac{1}{7} \right) + f \left( \frac{\frac{1}{7} + 1}{1 - 3 \cdot \frac{1}{7}} \right) = \frac{1}{7} \Rightarrow f \left( \frac{1}{7} \right) + f(2) = \frac{1}{7}   \] 4. Solve the system of equations:   We have the following system of equations:   \[     f(2) + f \left( -\frac{3}{5} \right) = 2 \quad \text{(1)} \ \]\[  f \left( -\frac{3}{5} \right) + f \left( \frac{1}{7} \right) = -\frac{3}{5} \quad \text{(2)} \  \]\[ f \left( \frac{1}{7} \right) + f(2) = \frac{1}{7} \quad \text{(3)}    \] 5. Add equations (1) and (3):  \[   (f(2) + f \left( -\frac{3}{5} \right)) + (f \left( \frac{1}{7} \right) + f(2)) = 2 + \frac{1}{7}   \]   \[   2f(2) + f \left( -\frac{3}{5} \right) + f \left( \frac{1}{7} \right) = \frac{15}{7}   \] 6. Substitute equation (2) into the result:   \[   2f(2) + \left( -\frac{3}{5} \right) = \frac{15}{7}   \]   \[   2f(2) - \frac{3}{5} = \frac{15}{7}   \] 7. Solve for \( f(2) \):   \[   2f(2) = \frac{15}{7} + \frac{3}{5}   \]   Find a common denominator:   \[   2f(2) = \frac{75}{35} + \frac{21}{35} = \frac{96}{35}   \]   \[   f(2) = \frac{48}{35}   \]Thus, the value of \( f(2) \) is \(\boxed{\frac{48}{35}}\). \textcolor{blue}{$\checkmark$}

\end{tcolorbox}

\begin{tcolorbox}[
  enhanced,
  breakable,
  fonttitle = \small\bfseries, 
  title=Competition mode (LLMs raise better solutions after interaction),
  colframe=Salmon!90!Black,   
  colback=Salmon!20,     
  boxrule=1pt,
  boxsep=2pt,
  left=5pt,
  right=5pt,
  fontupper=\footnotesize,
  halign title = flush center
]
\textbf{Problem:}

Find the minimum value of \(\frac{x^2 + 7}{\sqrt{x^2 + 3}}\) over all real numbers $x$.

\textbf{LLM1 (Idea Sharing):}

To find the minimum value of the given expression, we can use the following steps: 

Step 1: Rewrite the expression. We can rewrite the given expression as \[\frac{x^2 + 7}{\sqrt{x^2 + 3}} = \frac{x^2 + 3 + 4}{\sqrt{x^2 + 3}} = \frac{x^2 + 3}{\sqrt{x^2 + 3}} + \frac{4}{\sqrt{x^2 + 3}}\] Step 2: Simplify the expressionSimplifying the expression, we get\[\frac{x^2 + 3}{\sqrt{x^2 + 3}} + \frac{4}{\sqrt{x^2 + 3}} = \sqrt{x^2 + 3} + \frac{4}{\sqrt{x^2 + 3}}\] Step 3: Use the AM-GM inequalityUsing the AM-GM inequality, we have\[\sqrt{x^2 + 3} + \frac{4}{\sqrt{x^2 + 3}} \geq 2 \sqrt{\sqrt{x^2 + 3} \cdot \frac{4}{\sqrt{x^2 + 3}}} = 4\] Step 4: Find the equality condition. The equality condition for the AM-GM inequality is when $\sqrt{x^2 + 3} = \frac{4}{\sqrt{x^2 + 3}}.$ Solving for $x$, we get $x^2 + 3 = 4$, which gives $x = \pm 1$. 

Therefore, the final answer is: $\boxed{4}$. \textcolor{blue}{$\checkmark$}

\textbf{LLM2 (Idea Sharing):}

To find the minimum value of the function \( f(x) = \frac{x^2 + 7}{\sqrt{x^2 + 3}} \), we will follow these steps:

1. Define the function and its domain:   \[   f(x) = \frac{x^2 + 7}{\sqrt{x^2 + 3}}   \]   The domain of \( f(x) \) is all real numbers \( x \) because the expression under the square root, \( x^2 + 3 \), is always positive for all real \( x \).

2. Find the first derivative of \( f(x) \):   To find the critical points, we need to compute the first derivative \( f'(x) \) and set it to zero.   \[   f(x) = \frac{x^2 + 7}{\sqrt{x^2 + 3}}   \]  Let \( u = x^2 + 7 \) and \( v = \sqrt{x^2 + 3} \). Then \( f(x) = \frac{u}{v} \).   Using the quotient rule:  \[  f'(x) = \frac{u'v - uv'}{v^2}  \]  where \( u' = 2x \) and \( v' = \frac{x}{\sqrt{x^2 + 3}} \).   Substituting \( u \), \( u' \), \( v \), and \( v' \) into the quotient rule:  \[  f'(x) = \frac{(2x)(\sqrt{x^2 + 3}) - (x^2 + 7)\left(\frac{x}{\sqrt{x^2 + 3}}\right)}{(x^2 + 3)}  \]  Simplify the numerator:  \[  f'(x) = \frac{2x\sqrt{x^2 + 3} - \frac{x(x^2 + 7)}{\sqrt{x^2 + 3}}}{x^2 + 3} = \frac{x(x^2 - 1)}{(x^2 + 3)\sqrt{x^3 + 3}}  \]  Set the first derivative equal to zero to find critical points:  \[  \frac{x(x^2 - 1)}{(x^2 + 3)\sqrt{x^2 + 3}} = 0 \Rightarrow x = 0, \quad x = 1, \quad x = -1  \]

3. Evaluate \( f(x) \) at the critical points and endpoints:  \[  f(0) = \frac{0^2 + 7}{\sqrt{0^2 + 3}} = \frac{7}{\sqrt{3}} = \frac{7\sqrt{3}}{3}  \]  \[  f(1) = \frac{1^2 + 7}{\sqrt{1^2 + 3}} = \frac{8}{\sqrt{4}} = \frac{8}{2} = 4  \]  \[  f(-1) = \frac{(-1)^2 + 7}{\sqrt{(-1)^2 + 3}} = \frac{8}{\sqrt{4}} = \frac{8}{2} = 4  \]

4. Determine the minimum value:  Comparing the values:  \[  f(0) = \frac{7\sqrt{3}}{3} \approx 4.041  \]  \[  f(1) = 4  \]  \[  f(-1) = 4  \]  The minimum value is \( 4 \). Therefore, the minimum value of the function is \(\boxed{4}\). \textcolor{blue}{$\checkmark$}

\textbf{LLM1 (Idea Analysis):}

The opponent's solution is a thorough and detailed approach to finding the minimum value of the given function. Here's a critical analysis of his ideas: 

\textbf{Strengths}: 
1. Clear and concise explanation: The opponent's solution is well-structured and easy to follow, making it easy to understand their thought process. 2. Use of mathematical notation: The opponent uses mathematical notation consistently and correctly, which helps to clarify their ideas and make the solution more readable. 3. Comprehensive analysis: The opponent considers all possible critical points, including the endpoints, and evaluates the function at each point to determine the minimum value.

\textbf{Weaknesses}: 1. Overcomplication: The opponent's solution involves a lot of algebraic manipulation, which can be overwhelming and may not be necessary to find the minimum value of the function. 2. Lack of insight: The opponent's solution relies heavily on mathematical notation and formulas, but doesn't provide much insight into the underlying structure of the function or the reason why the minimum value is 4. 3. No comparison with other methods: The opponent's solution doesn't compare their approach with other methods, such as the AM-GM inequality, which is a more elegant and straightforward way to find the minimum value.

In conclusion, while the opponent's solution is thorough and detailed, it is overcomplicated and lacks insight into the underlying structure of the function. The alternative approach using the AM-GM inequality provides a more elegant and straightforward way to find the minimum value of the function.

\textbf{LLM2 (Idea Analysis):}

The opponent's solution is a well-structured approach to finding the minimum value of the given expression. Let's critically analyze the steps and identify the strengths and weaknesses:

\textbf{Strengths}: 1. Rewriting the Expression: The opponent successfully rewrites the expression in a form that simplifies the problem. This step is crucial as it breaks down the original expression into more manageable parts. 2. Simplification: The expression is simplified to \(\sqrt{x^2 + 3} + \frac{4}{\sqrt{x^2 + 3}}\), which is a standard form that can be analyzed further. 3. Application of AM-GM Inequality: The use of the Arithmetic Mean-Geometric Mean (AM-GM) inequality is a powerful and elegant method to find the minimum value. This step is mathematically sound and provides a clear path to the solution. 

\textbf{Weaknesses}: 1. Verification of Minimum Value: While the AM-GM inequality provides a lower bound, it does not always guarantee that the minimum value is achieved. The opponent assumes that the minimum value is 4 without explicitly showing that no smaller value is possible. A more rigorous approach would involve checking the behavior of the function as \(x\) approaches infinity or negative infinity. 2. Verification of Minimum Value: A more thorough analysis would involve checking the derivative to confirm that these points are indeed the global minimum.

Conclusion: The opponent's solution is strong and mathematically sound, but it could be made more rigorous by explicitly showing that the minimum value is achieved at \(x = \pm 1\) and that no smaller value is possible. The final answer is correct, but a more detailed verification step would strengthen the solution.

\textbf{LLM1 (Idea Fusion):}

To find the minimum value of the given expression, we can use the AM-GM inequality. 

Step 1: Apply the AM-GM inequality. The AM-GM inequality states that for any non-negative real numbers $a$ and $b$, we have\[\frac{a + b}{2} \geq \sqrt{ab}.\] We can rewrite the given expression as\[\frac{x^2 + 7}{\sqrt{x^2 + 3}} = \frac{x^2 + 3 + 4}{\sqrt{x^2 + 3}} = \sqrt{x^2 + 3} + \frac{4}{\sqrt{x^2 + 3}}.\] 

Step 2: Use the AM-GM inequality on the expression. Applying the AM-GM inequality to the expression, we get\[\sqrt{x^2 + 3} + \frac{4}{\sqrt{x^2 + 3}} \geq 2\sqrt{\sqrt{x^2 + 3} \cdot \frac{4}{\sqrt{x^2 + 3}}} = 4.\] 

Step 3: Find the equality condition. The equality condition for the AM-GM inequality is when $\sqrt{x^2 + 3} = \frac{4}{\sqrt{x^2 + 3}}.$ Solving for $x$, we get $x^2 + 3 = 4$, which gives $x = \pm 1$. 

Step 4: Check the equality condition. We need to check if the equality condition is satisfied for $x = \pm 1$. Substituting $x = \pm 1$ into the original expression, we get\[\frac{(\pm 1)^2 + 7}{\sqrt{(\pm 1)^2 + 3}} = \frac{8}{2} = 4.\] 

Step 5: Conclusion. Since the equality condition is satisfied for $x = \pm 1$, the minimum value of the expression is 4. The final answer is: $\boxed{4}$ \textcolor{blue}{$\checkmark$}

\textbf{LLM2 (Idea Fusion):}

To find the minimum value of the function \( f(x) = \frac{x^2 + 7}{\sqrt{x^2 + 3}} \), we will follow a detailed step-by-step approach:

1. Rewrite the Expression:   \[   f(x) = \frac{x^2 + 7}{\sqrt{x^2 + 3}} = \sqrt{x^2 + 3} + \frac{4}{\sqrt{x^2 + 3}}   \]

2. Simplify the Expression:   We have simplified the expression to:   \[  f(x) = \sqrt{x^2 + 3} + \frac{4}{\sqrt{x^2 + 3}}  \]

3. Use the AM-GM Inequality:  The Arithmetic Mean-Geometric Mean (AM-GM) inequality states that for any non-negative real numbers \(a\) and \(b\),  \[  \frac{a + b}{2} \geq \sqrt{ab}  \]  Applying this to our expression:  \[  \sqrt{x^2 + 3} + \frac{4}{\sqrt{x^2 + 3}} \geq 2 \sqrt{\sqrt{x^2 + 3} \cdot \frac{4}{\sqrt{x^2 + 3}}} = 4  \]  Therefore,  \[  \sqrt{x^2 + 3} + \frac{4}{\sqrt{x^2 + 3}} \geq 4  \]

4. Find the Equality Condition:  The equality in the AM-GM inequality holds when:  \[  \sqrt{x^2 + 3} = \frac{4}{\sqrt{x^2 + 3}}  \]  Solving for \(x\):  \[  (\sqrt{x^2 + 3})^2 = \left(\frac{4}{\sqrt{x^2 + 3}}\right)^2  \]  \[  x^2 + 3 = \frac{16}{x^2 + 3}  \]  Let \( y = \sqrt{x^2 + 3} \). Then:  \[  y^2 = \frac{16}{y^2}  \]  \[  y^4 = 16  \]  \[  y^2 = 4 \quad \text{(since \(y \geq 0\))}  \]  \[  y = 2  \]  Therefore:  \[  \sqrt{x^2 + 3} = 2  \]  \[  x^2 + 3 = 4  \]  \[  x^2 = 1  \]  \[  x = \pm 1  \]

5. Verify the Minimum Value:  Substitute \(x = \pm 1\) into the original function to verify:  \[  f(1) = \frac{1^2 + 7}{\sqrt{1^2 + 3}} = \frac{8}{2} = 4  \]  \[  f(-1) = \frac{(-1)^2 + 7}{\sqrt{(-1)^2 + 3}} = \frac{8}{2} = 4  \]

6. Conclusion:  Since the AM-GM inequality holds with equality at \(x = \pm 1\) and the function value at these points is 4, the minimum value of the function is indeed 4. The final answer is: $\boxed{4}$ 
\textcolor{blue}{$\checkmark$}
\end{tcolorbox}




\end{document}